\documentclass[10pt,twocolumn,letterpaper]{article}
\usepackage[pagenumbers]{cvpr}
\usepackage{graphicx}
\usepackage{amsmath,amssymb}
\usepackage{booktabs,multirow,tabularx,colortbl}
\usepackage{microtype}
\usepackage{placeins}
\usepackage{enumitem}
\usepackage{xspace}
\usepackage{tikz}
\usetikzlibrary{positioning,arrows.meta,calc,fit,backgrounds}
\usepackage{pgfplots}
\usepackage{algorithm}
\usepackage[noend]{algpseudocode}
\pgfplotsset{compat=1.18}
\definecolor{accent}{HTML}{39757C}
\definecolor{upcol}{HTML}{3478E6}
\definecolor{downcol}{HTML}{D9822B}
\definecolor{rejcol}{HTML}{C0392B}
\definecolor{acccol}{HTML}{27AE60}
\definecolor{inkgray}{HTML}{4A4A4A}
\definecolor{pendgray}{HTML}{8C8C8C}
\definecolor{rowgray}{gray}{0.93}
\definecolor{cvprblue}{rgb}{0.21,0.49,0.74}
\usepackage[pagebackref,breaklinks,colorlinks,allcolors=cvprblue]{hyperref}
\usepackage[capitalize]{cleveref}
\newcommand{\method}{AffordJEV\xspace}

\newcommand{\smallhead}[1]{\noindent\textbf{#1}\ }
\graphicspath{{figures/pdf/}{figures/assets/}}
\hypersetup{pdftitle={AffordJEV: Affordance Grounding via Hierarchical Verification},pdfauthor={Jincheng Li; Yutian Zhang; Jianyu Zhang; Xinshun Wang; Jun Liu; Junsong Yuan; Mengyuan Liu},pdfsubject={arXiv preprint}}

\title{AffordJEV: Affordance Grounding via Hierarchical Verification}
\author{%
  Jincheng Li$^{1}$\thanks{Equal contribution.}\quad
  Yutian Zhang$^{1}$\footnotemark[1]\quad
  Jianyu Zhang$^{1}$\quad
  Xinshun Wang$^{1}$\\
  Jun Liu$^{2}$\quad
  Junsong Yuan$^{3}$\quad
  Mengyuan Liu$^{1}$\thanks{Corresponding author.}\\[2pt]
  $^{1}$Peking University, China\quad
  $^{2}$Lancaster University, UK\\
  $^{3}$State University of New York at Buffalo, USA
}
\begin{document}
\maketitle
\suppressfloats[t]
\begin{figure}[t]
  \centering
  \includegraphics[width=\linewidth]{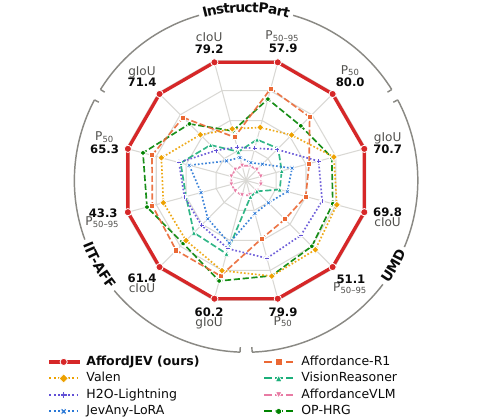}
  \caption{\textbf{Results on InstructPart, UMD and IIT-AFF} (\%). Each axis is one metric on a convex scale over
  the eight methods; the numbers are our scores (\cref{tab:instructpart,tab:umd,tab:iit}). Dashed: methods with
  their own inference paths; dotted: System 1 judges in our pipeline.}
  \label{fig:teaser}
\end{figure}
\begin{abstract}
Affordance grounding asks which visible part of an object performs a requested operation, a basic capability for
embodied manipulation and human--robot interaction. Most existing methods follow a System 2 paradigm: a
multimodal language model first writes a chain of thought and then turns its conclusion into a box or points for a
segmenter. This incurs considerable inference cost, gives the semantic decision little direct control over the
extent of the final region, and makes mistakes hard to correct. Yet locating an affordance is essentially a
perceptual judgment that need not be argued out in words. We therefore propose \method, in which frozen segmenters
propose candidate regions and a judge decides directly which of them perform the operation, reading each decision
from a single forward pass without generating any textual explanation, in the manner of a System 1 model. Our
contributions are twofold. (1) An inference framework that arranges these single-pass decisions in a fixed
hierarchy, verifying candidate regions from object to part to sub-part, with evidence passed between the levels in
both directions so that the levels correct one another. (2) A judge model that learns from the requests this
framework produces, judging all candidates of a level in a single forward pass, with each candidate read from the
pixels it covers. Trained only on InstructPart and evaluated with definition-aligned operation
words, \method outperforms chain-of-thought methods and released general-purpose judges run in the same framework, both in
domain and in zero-shot transfer to UMD and IIT-AFF, while issuing a fraction of their judge requests. To our
knowledge, \method is the first approach to bring System 1 models to affordance grounding.
\end{abstract}

\section{Introduction}
\label{sec:intro}
Affordance grounding links an object and an intended operation to the visible region that performs it: a robot asked
to cut with a knife needs the blade, not merely a correctly recognized knife. This capability underlies embodied
manipulation and human--robot interaction, where knowing what an object is matters less than knowing where to act on
it. The output must be spatially precise, but the difficulty is mostly semantic: a handle, a working surface and the
support under an object are all clean segments, and only one of them serves the requested operation.

The prevailing recipe lets a multimodal language model reason before it answers. Affordance-R1~\cite{wang2026affordancer1}
and VisionReasoner~\cite{liu2026visionreasoner} write a reasoning trace, decode a box or points and hand them to a
segmenter, and OP-HRG~\cite{mehrab2026reasoning} adds a self-check before segmentation. This is a System 2 procedure in
the sense of Kahneman~\cite{kahneman2011thinking}: slow, deliberate and general. Its conclusion reaches the image only
through a box or a few points, so the segmenter, not the reasoning, decides how much of the object is returned, and a
wrong conclusion leaves nothing that could correct it. Yet grounding an affordance rarely calls for verbal
deliberation, especially for familiar everyday objects. Finding the handle of a mug does take steps, first the mug and
then its parts, but perceptual ones: vision takes in global structure before local detail~\cite{navon1977forest},
recognizes objects by their parts~\cite{biederman1987recognition} and chains such elementary operations into visual
routines~\cite{ullman1984visual}, all without words. This raises a question: does locating the part that performs an
operation require writing out an argument at all?

In this paper we present \method, which grounds affordances in the manner of a System 1 model. Frozen segmenters
propose candidate regions, and a judge, a multimodal model that scores each candidate region, decides directly which of
them perform the operation, reading every score from a single forward pass instead of decoding text. Our key insight is
that the reasoning the task needs can live in the structure of the questions rather than in generated text. \method
rests on two components designed together. Because the judge returns a probability for every candidate rather than
one decoded answer, its decisions can be combined within a level, compared across levels and revised, which is what
the framework exploits.

The first, a \emph{hierarchical verification framework}, decides which candidates the judge scores, and in what order. One
question over every region of an image is fragile: the queried part is small, regions of other objects compete with it,
and a wrong object is never corrected. People narrow down instead, and they revise: if no part of the chosen object fits
the operation, the object was probably wrong. The framework turns this into a hierarchy of region questions. The judge
verifies the object, selects its parts and refines them with sub-parts, and its verdicts pass between the levels: weak
part evidence under an unsure object reopens the object decision, and sub-part verdicts trim an accepted part or rescue
a rejected one. Every step asks a question of the same form --- which of these regions answer the query --- so one judge
serves all levels. \method is thus not reasoning-free, but its reasoning is not a chain of thought: the order of the
questions is fixed in advance rather than chosen by the model, and no step writes intermediate content. OP-HRG decomposes the task similarly, first the object and then its
part; what separates the two is how each step is decided: by a written argument or by one forward pass.

The second component is a \emph{joint region judge} built for these questions. Released System 1
judges~\cite{valen2026,h2o2026lightning,simplejev2026jevany} are general-purpose decision models: they see one candidate
per request, learn its identity from a marker painted onto the image, pool the whole context into a single vector in
which a small part is diluted, and score through a head trained from scratch on large collections of unrelated
decisions. Yet on the regions our segmenters propose, accepting exactly the candidates that lie mostly inside the ground
truth would clearly exceed every method we evaluate, so the bottleneck is the judge rather than the candidates. In our
judge, each candidate becomes one token pooled from the vision encoder's own output over its mask, so the judge reads
the pixels a candidate covers instead of a drawn marker. All candidates of a level share one sequence and one forward
pass, and a joint decision head lets them see one another, because the framework outputs the union of what it accepts. The
choice between object hypotheses is scored by a verbalizer, which reads the LLM's own \textsf{Yes}/\textsf{No} answer
tokens and thereby reuses the answer interface of the pretrained Qwen3.5-4B~\cite{qwen2026} instead of a head trained
from scratch. The judge learns from the requests the framework itself produces on the training photos.

Comparing across datasets also needs care: the same operation word names different regions in different datasets ---
InstructPart marks a vessel's body for \emph{grasp}, UMD and IIT-AFF a tool's handle --- so we align the conflicting
words to each dataset's annotation definitions through a mapping fixed in advance and applied to every method. As shown
in \cref{fig:teaser}, trained only on InstructPart, \method outperforms chain-of-thought methods and released System 1
judges run in the same framework, both in domain and in zero-shot transfer to UMD~\cite{myers2015umd} and
IIT-AFF~\cite{nguyen2017iit}, while decoding no text and issuing far fewer judge requests than per-candidate judges.

To summarize, our contributions are as follows:
\begin{itemize}[leftmargin=*,itemsep=1pt,topsep=3pt]
\item We propose a \textbf{hierarchical verification framework} that grounds affordances by verifying candidate
regions level by level --- object, part, sub-part --- and passes evidence between the levels in both directions, so
that one judge answering one form of question serves the whole hierarchy.
\item We introduce a \textbf{joint region judge}, a System 1 model trained from the base weights on the framework's own
requests, which reads each candidate from the pixels it covers, judges all candidates of a level in one forward pass,
and scores object choices with a verbalizer on the model's own answer tokens.
\item We evaluate on InstructPart, UMD and IIT-AFF under a definition-aligned protocol, where \method outperforms all
published baselines in domain and in zero-shot transfer, and we ablate every component of the framework and the judge.
\end{itemize}

\section{Related Work}
\label{sec:related}
\smallhead{Affordance grounding.}
Affordances describe what an object offers to an agent~\cite{gibson1979ecological}. Early supervised models segment
them per pixel~\cite{do2018affordancenet} on benchmarks that pair tool parts with functional labels, such as
UMD~\cite{myers2015umd} and IIT-AFF~\cite{nguyen2017iit}; later benchmarks cover hand-occluded
containers~\cite{apicella2023affordance}, interaction heatmaps~\cite{luo2022agd} and part-level masks for task
instructions in everyday photos~\cite{wan2025instructpart}. Weaker supervision comes from human interaction in videos
and images~\cite{nagarajan2019hotspots,li2023locate,mo2021where2act}, and OOAL adapts foundation features from sparse
labels~\cite{li2024ooal}, whose geometric and interaction cues recent work probes~\cite{zhang2026probing}. With
multimodal language models, AffordanceLLM brings vision--language knowledge to the problem~\cite{qian2024affordancellm},
AffordanceVLM couples such a model with a mask decoder for reasoning-based affordance segmentation~\cite{wu2025ragnet},
Affordance-R1 trains reasoning-based grounding with reinforcement learning~\cite{wang2026affordancer1}, and A4-Agent
decouples reasoning from grounding by coordinating generative, vision--language and segmentation models without
training~\cite{zhang2025a4agent}; AFUN explores a foundation-model formulation~\cite{wang2026afun}, and ThinkAfford
reasons about affordances in cluttered 3D scenes~\cite{lin2026thinkafford}. We train on InstructPart, transfer without
fine-tuning to UMD and IIT-AFF, and make direct decisions on proposed regions instead of reasoning in text. Since each
decision selects whole proposed regions rather than predicting a dense map, the output follows the boundaries of the
parts the segmenters find.

\smallhead{Reasoning-based segmentation.}
Promptable segmenters supply transferable masks~\cite{kirillov2023sam,ravi2024sam2}, SAM~3 segments all instances of a
named concept~\cite{carion2025sam3}, and part segmenters and
datasets~\cite{li2024partglee,he2022partimagenet,ramanathan2023paco,sun2023vlpart} define parts by name, not by the
operation they serve. Reasoning segmentation lets a multimodal model decide what to segment: LISA couples a language
model with a mask decoder~\cite{lai2024lisa}, Set-of-Mark prompting lets it refer to marked regions~\cite{yang2023som},
and AffordanceSAM connects promptable segmentation to affordances~\cite{jiang2025affordancesam}. Chain-of-thought
prompting~\cite{wei2022cot} and reinforcement learning on verifiable rewards~\cite{guo2025deepseekr1} let such models
reason before answering: VisionReasoner unifies several perception tasks behind a reasoning
stage~\cite{liu2026visionreasoner}, and OP-HRG hierarchically localizes the object and its part, then self-checks its
answer before segmentation~\cite{mehrab2026reasoning}. Visual programming composes vision modules through programs that a
language model writes for each query~\cite{gupta2023visprog,suris2023vipergpt}. In all of these, the model produces the
steps as text or code. \method fixes its steps in advance, as a hierarchy of region questions in which later decisions
can revise earlier ones, and decides each step in one forward pass.

\smallhead{Multimodal models as judges.}
Another line uses models as judges of candidate answers rather than as generators~\cite{zheng2023judging,lightman2024verify}.
Valen exposes choice and yes/no decisions of a multimodal backbone through a role-based scoring head~\cite{valen2026},
and H2O-Lightning and JevAny offer similar direct interfaces~\cite{h2o2026lightning,simplejev2026jevany}. These are
general-purpose decision models: one candidate per request, candidate identity supplied by a drawn marker, and a head
trained on large collections of unrelated decisions. We refer to models that answer in one forward pass without
generating text as System 1 models; the term describes how each decision is made, not whether the task is decomposed.
Our judge keeps the direct decision and changes its interface. Candidates enter as features pooled over their masks,
following region features in detection and segmentation~\cite{he2017maskrcnn,kirillov2023sam} and region-level
multimodal models~\cite{zhang2023gpt4roi,you2024ferret,yuan2024osprey}; object choices are read through a verbalizer,
the answer token an autoregressive model would have produced, as in prompt-based
classification~\cite{schick2021exploiting,brown2020language}; and the regions of one level attend to one another, as in
set prediction for detection~\cite{carion2020detr}, though there the set removes duplicates rather than composing an
answer.

\section{Method}
\label{sec:method}

\definecolor{lmcol}{HTML}{6A55B4}
\definecolor{levcol}{HTML}{2E4A62}
\definecolor{rc1}{RGB}{196,78,160}
\definecolor{rc2}{RGB}{232,140,40}
\definecolor{rc6}{RGB}{205,165,40}
\definecolor{objcol}{RGB}{52,120,230}%
\providecommand{\figice}{\raisebox{-0.18ex}{\includegraphics[height=1.55ex]{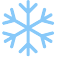}}}
\providecommand{\figfire}{\raisebox{-0.18ex}{\includegraphics[height=1.55ex]{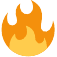}}}
\begin{figure*}[t]
\centering
\begin{tikzpicture}[x=1cm, y=1cm, font=\sffamily\fontsize{6.5}{7.5}\selectfont,
  pan/.style={inner sep=0, draw=inkgray!30, line width=0.4pt},
  num/.style={font=\sffamily\fontsize{6.5}{7.5}\selectfont, text=inkgray, inner sep=1pt},
  lab/.style={font=\sffamily\fontsize{6.6}{7.6}\selectfont, text=inkgray, inner sep=1pt},
  ptab/.style={font=\sffamily\fontsize{7.6}{8.6}\selectfont\bfseries, fill=white, inner xsep=3pt, inner ysep=1pt, anchor=west},
  card/.style={rounded corners=3pt, draw=levcol!35, fill=levcol!3, line width=0.6pt},
  badge/.style={circle, fill=levcol, text=white, inner sep=0pt, minimum size=0.30cm,
                font=\sffamily\fontsize{6.4}{7}\selectfont\bfseries},
  lbadge/.style={circle, fill=levcol, text=white, inner sep=0pt, minimum size=0.24cm,
                font=\sffamily\fontsize{6}{6.4}\selectfont\bfseries},
  hname/.style={font=\sffamily\fontsize{7.2}{8}\selectfont\bfseries, text=levcol, anchor=west, inner sep=0pt},
  flow/.style={-{Stealth[length=1.8mm,width=1.6mm]}, line width=0.9pt, draw=inkgray!60},
  wire/.style={-{Stealth[length=1.2mm,width=1.0mm]}, draw=inkgray!60, line width=0.45pt},
  tok/.style={minimum width=0.20cm, minimum height=0.22cm, inner sep=0, line width=0.3pt, rounded corners=0.6pt},
  lbox/.style={rounded corners=2pt, draw=lmcol!75, fill=lmcol!9, line width=0.6pt},
  abox/.style={rounded corners=2pt, draw=accent!85, fill=accent!9, line width=0.6pt},
  qchip/.style={rounded corners=1.2pt, draw=inkgray!55, fill=white, line width=0.4pt, inner xsep=1.3pt, inner ysep=0pt,
                minimum height=0.28cm, font=\sffamily\fontsize{6.4}{7.4}\selectfont, text=inkgray},
  tchip/.style={qchip, dashed, dash pattern=on 1.2pt off 0.8pt, fill=gray!7},
  rk/.style={tok, minimum width=0.30cm, minimum height=0.34cm, draw=none, text=white, inner xsep=0.5pt,
             font=\sffamily\fontsize{6.4}{7.4}\selectfont},
  mname/.style={font=\sffamily\fontsize{6.8}{7.8}\selectfont\bfseries, align=center, inner sep=1pt},
  vmark/.style={font=\sffamily\bfseries\fontsize{7}{7}\selectfont, inner sep=0},
  vdot/.style={circle, inner sep=0pt, minimum size=0.13cm, draw=white, line width=0.4pt},
  tag/.style={fill=levcol, text=white, rounded corners=1.6pt, inner xsep=2.6pt, inner ysep=1.1pt,
              font=\sffamily\fontsize{6.6}{7.6}\selectfont\bfseries},
]
\newcommand{\vyes}[3]{\node[inner sep=0, draw=acccol, line width=0.9pt] at (#1,#2) {#3};
  \node[circle, fill=acccol, draw=white, line width=0.5pt, inner sep=0, minimum size=0.23cm, text=white,
        font=\fontsize{6.4}{6.4}\selectfont\bfseries] at (#1+0.25,#2+0.25) {\checkmark};}
\newcommand{\vno}[3]{\node[inner sep=0, draw=rejcol, line width=0.9pt] at (#1,#2) {#3};
  \node[circle, fill=rejcol, draw=white, line width=0.5pt, inner sep=0, minimum size=0.23cm, text=white,
        font=\fontsize{6.8}{6.8}\selectfont\bfseries] at (#1+0.25,#2+0.25) {$\times$};}
\newcommand{\meter}[3]{%
  \fill[rejcol!8] (#1-0.24,#2) rectangle (#1+0.24,#2+#3/2);
  \fill[acccol!12] (#1-0.24,#2+#3/2) rectangle (#1+0.24,#2+#3);
  \draw[draw=inkgray!30, line width=0.35pt] (#1-0.24,#2) rectangle (#1+0.24,#2+#3);
  \draw[draw=inkgray!75, line width=0.55pt, dash pattern=on 1.4pt off 0.9pt] (#1-0.24,#2+#3/2) -- (#1+0.24,#2+#3/2);
  \node[vmark, text=acccol] at (#1,#2+#3+0.15) {\checkmark};
  \node[vmark, text=rejcol] at (#1,#2-0.14) {$\times$};
}
\def\aY{0.10}%
\def\aS{0.80}%
\begin{scope}[on background layer]
  \draw[rounded corners=4pt, draw=levcol!30, fill=white, line width=0.6pt] (0.00,4.68) rectangle (17.40,{9.03+\aY});
\end{scope}
\node[ptab, text=levcol] at (0.30,{9.03+\aY}) {(a) Hierarchical Verification Framework};

\begin{scope}[yshift=\aY cm]
\node[font=\sffamily\fontsize{6.2}{7}\selectfont, text=inkgray, anchor=east] at (1.10,8.45) {object};
\node[rounded corners=2pt, draw=objcol, fill=objcol!8, line width=0.7pt, inner xsep=2.4pt, inner ysep=1.4pt, anchor=west,
      font=\sffamily\bfseries\fontsize{7.2}{8.2}\selectfont, text=objcol!70!black] at (1.16,8.45) {suitcase};
\node[font=\sffamily\fontsize{6.2}{7}\selectfont, text=inkgray, anchor=east] at (1.10,8.03) {operation};
\node[rounded corners=2pt, draw=acccol, fill=acccol!8, line width=0.7pt, inner xsep=2.4pt, inner ysep=1.4pt, anchor=west,
      font=\sffamily\bfseries\fontsize{7.2}{8.2}\selectfont, text=acccol!60!black] at (1.16,8.03) {drag};
\node[pan, anchor=north] (in) at (1.30,7.72) {\includegraphics[width=1.60cm]{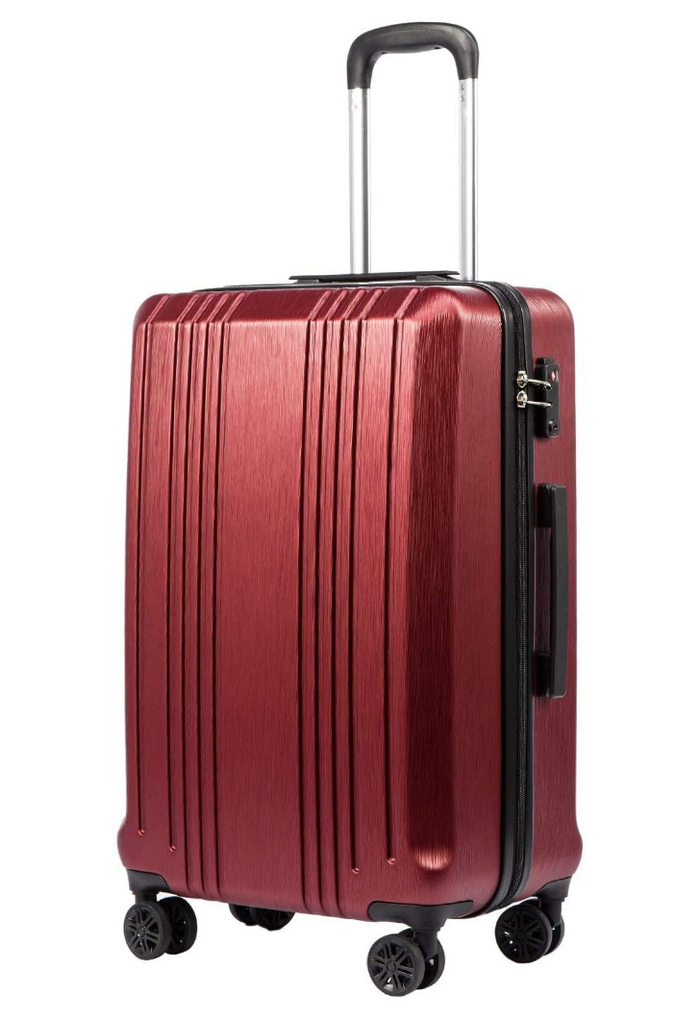}};

\begin{scope}[xshift=\aS cm]
\foreach \xa/\xb/\n/\name/\sh/\bd in {1.70/4.75/1/Object/60/40, 6.125/9.225/2/Part/80/55, 10.60/13.85/3/Sub-part/100/70} {
  \fill[levcol!3, rounded corners=3pt] (\xa,5.02) rectangle (\xb,8.72);
  \begin{scope}
    \clip[rounded corners=3pt] (\xa,5.02) rectangle (\xb,8.72);
    \fill[levcol!\sh] (\xa,8.30) rectangle (\xb,8.72);
  \end{scope}
  \draw[rounded corners=3pt, draw=levcol!\bd, line width=0.75pt] (\xa,5.02) rectangle (\xb,8.72);
  \node[circle, fill=white, text=levcol!\sh, inner sep=0pt, minimum size=0.29cm,
        font=\sffamily\fontsize{6.6}{7}\selectfont\bfseries] at (\xa+0.27,8.51) {\n};
  \node[font=\sffamily\fontsize{8}{9}\selectfont\bfseries, text=white, anchor=west, inner sep=0] at (\xa+0.48,8.51) {\name};
}

\coordinate (zA1) at (4.68,7.03); \coordinate (zA0) at (4.68,6.43);     %
\coordinate (zB1) at (6.175,8.26); \coordinate (zB0) at (6.175,5.20);     %
\coordinate (zC1) at (8.955,8.08); \coordinate (zC0) at (8.955,7.48);     %
\coordinate (zD1) at (10.65,8.254); \coordinate (zD0) at (10.65,5.206); %
\tikzset{zoomline/.style={draw=levcol!40, line width=0.45pt}}
\shade[left color=levcol!17, right color=levcol!3] (zA1) -- (zB1) -- (zB0) -- (zA0) -- cycle;
\shade[left color=levcol!17, right color=levcol!3] (zC1) -- (zD1) -- (zD0) -- (zC0) -- cycle;
\draw[zoomline] (zA1) -- (zB1); \draw[zoomline] (zA0) -- (zB0);
\draw[zoomline] (zC1) -- (zD1); \draw[zoomline] (zC0) -- (zD0);

\node[pan, anchor=south west] (obj) at (1.90,5.24) {\includegraphics[width=1.95cm]{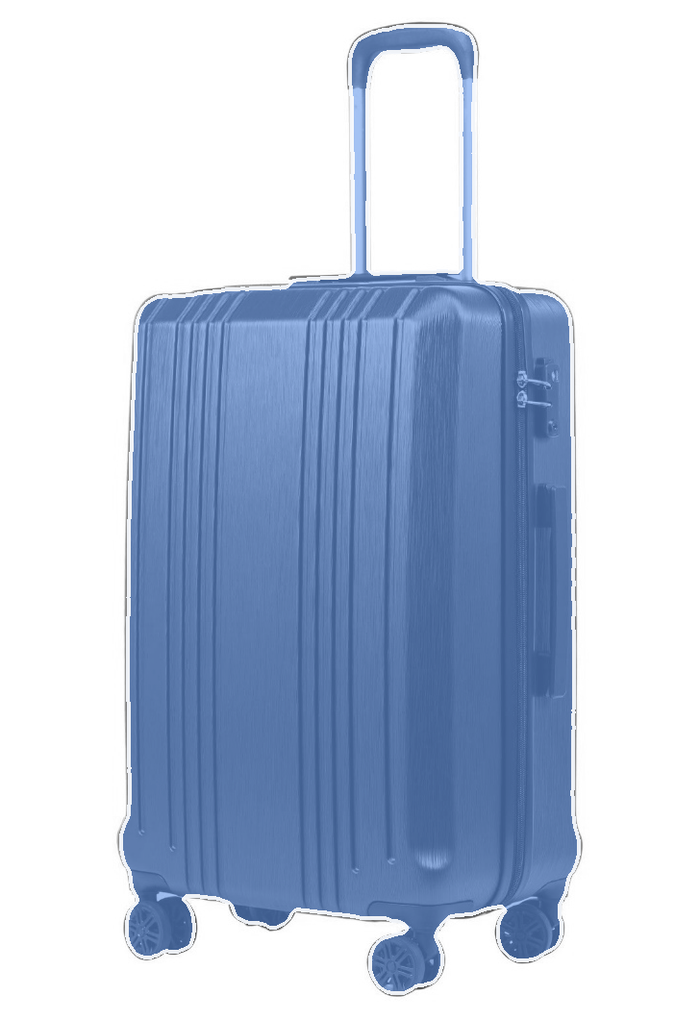}};
\node[fill=levcol, text=white, inner xsep=1.6pt, inner ysep=0.6pt, rounded corners=1pt, anchor=north west,
      font=\sffamily\fontsize{6.4}{7.4}\selectfont] at ([shift={(1pt,-1pt)}]obj.north west) {$F_0$};

\node[pan] (parts) at (7.225,6.73) {\includegraphics[width=2.10cm]{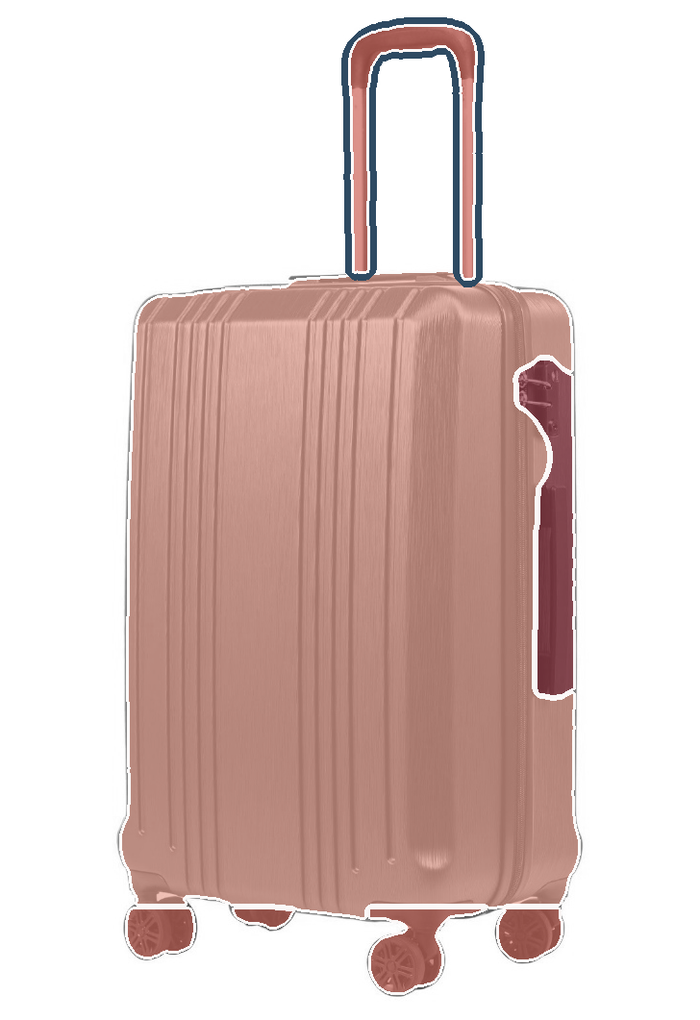}};

\node[pan] (sub) at (11.85,6.73) {\includegraphics[width=2.40cm]{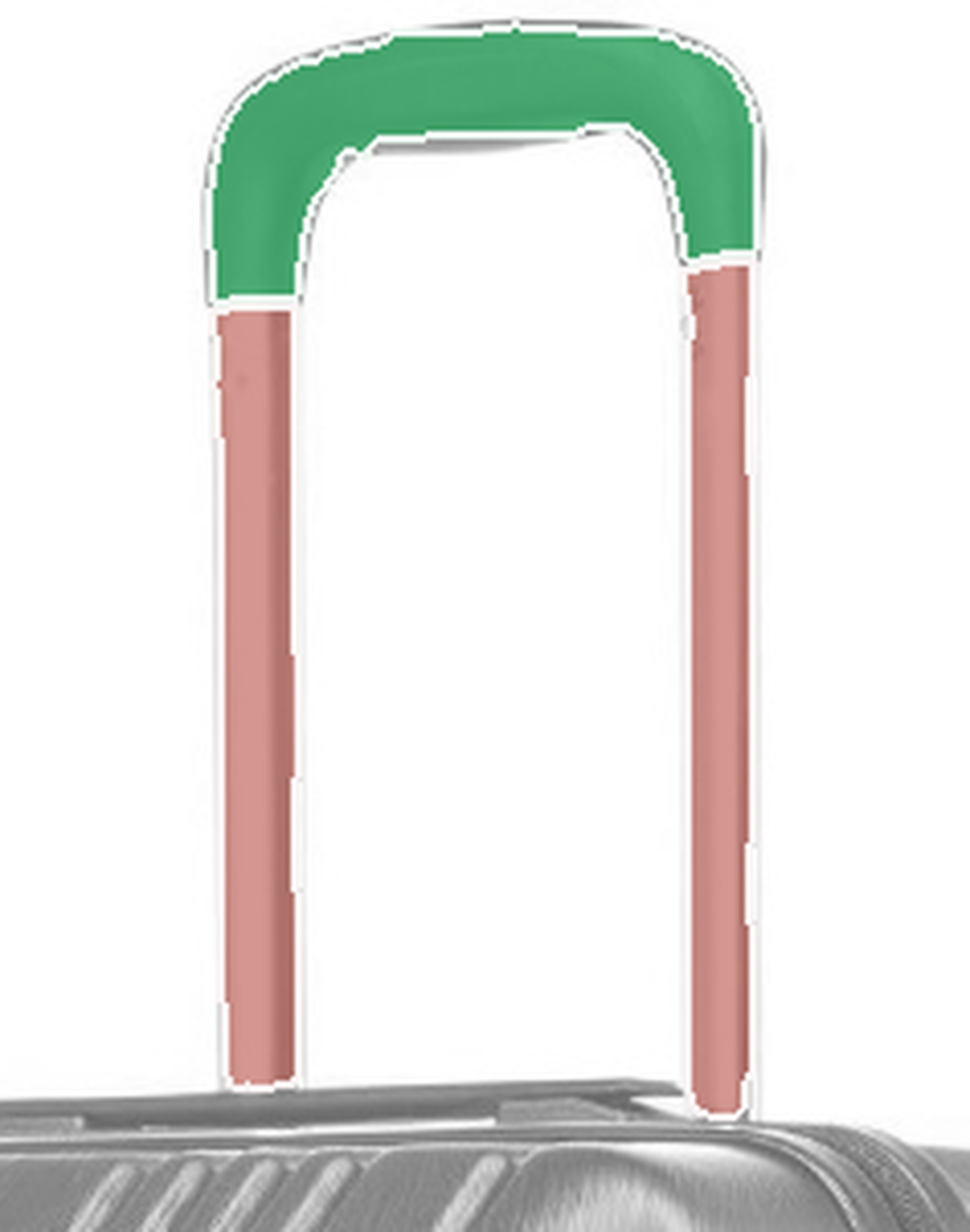}};

\vyes{4.35}{6.73}{\includegraphics[width=0.60cm]{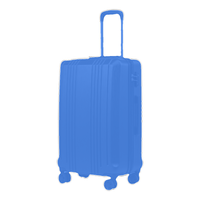}}
\vno{8.625}{7.78}{\includegraphics[width=0.60cm]{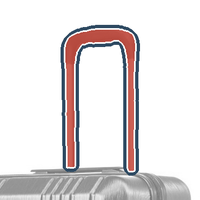}}
\vno{8.625}{7.08}{\includegraphics[width=0.60cm]{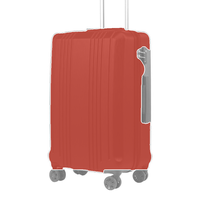}}
\vno{8.625}{6.38}{\includegraphics[width=0.60cm]{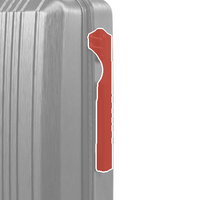}}
\vno{8.625}{5.68}{\includegraphics[width=0.60cm]{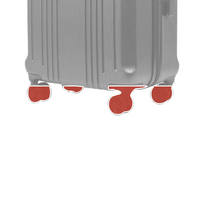}}
\vno{13.40}{7.43}{\includegraphics[width=0.60cm]{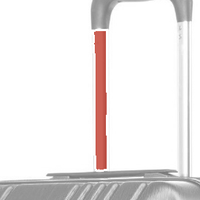}}
\vno{13.40}{6.73}{\includegraphics[width=0.60cm]{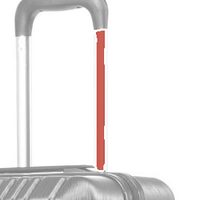}}
\vyes{13.40}{6.03}{\includegraphics[width=0.60cm]{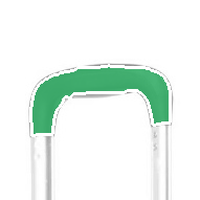}}

\end{scope}

\node[lab, anchor=south] at (16.10,8.301) {output};
\node[pan] (out) at (16.10,6.73) {\includegraphics[width=2.10cm]{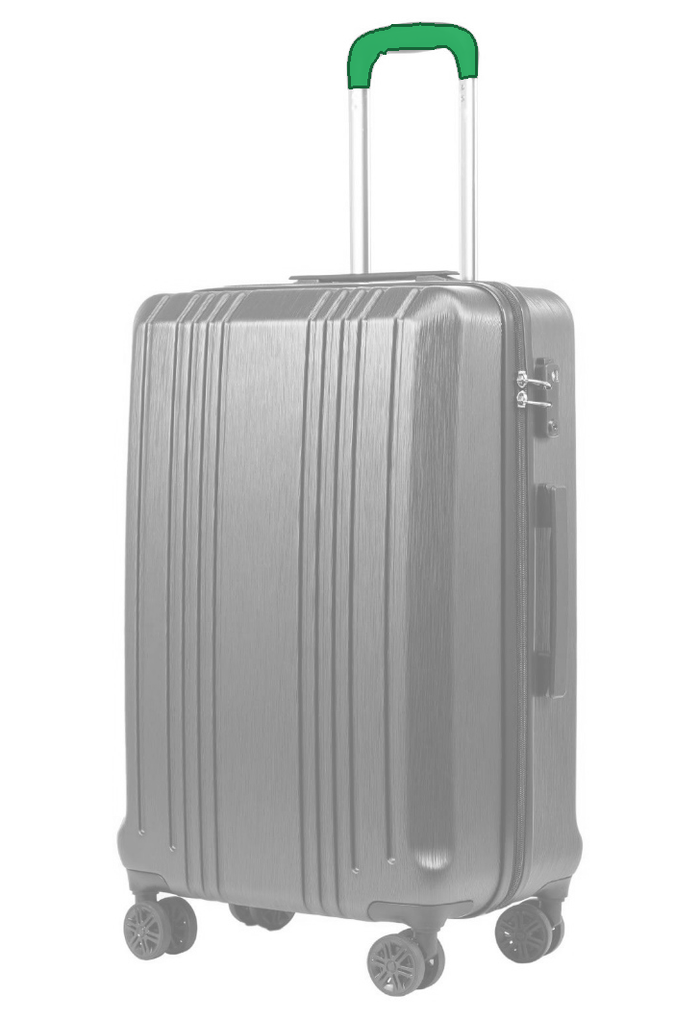}};

\draw[flow] (in.east |- 0,6.73) -- ({1.70+\aS},6.73);
\draw[flow] ({13.85+\aS},6.73) -- (15.02,6.73);
\begin{scope}[xshift=\aS cm]
\tikzset{move/.style={-{Stealth[length=2.1mm,width=1.9mm]}, draw=levcol, line width=1.3pt},
         mtag/.style={tag, fill=levcol, font=\sffamily\fontsize{6.6}{7.6}\selectfont\bfseries, inner xsep=2.4pt, inner ysep=1.2pt}}
\draw[move] (9.225,6.73) -- (10.60,6.73);
\node[mtag] at (9.9125,7.04) {split};
\draw[move, dash pattern=on 3.6pt off 2.0pt] (6.125,6.73) -- (4.68,6.73);
\node[mtag] at (5.4375,7.04) {backtrack};
\end{scope}
\end{scope}

\begin{scope}[xshift=\aS cm]
\draw[-{Stealth[length=1.6mm,width=1.5mm]}, draw=lmcol!80, line width=0.8pt, dashed, dash pattern=on 1.4pt off 1pt]
  (4.35,4.10) -- (4.35,{6.37+\aY});
\node[font=\sffamily\fontsize{6.8}{7.8}\selectfont\bfseries, text=lmcol, anchor=west, inner sep=0] at (4.45,4.92) {+\,Verbalizer};
\draw[-{Stealth[length=1.6mm,width=1.5mm]}, draw=accent, line width=0.8pt] (8.625,4.10) -- (8.625,{5.32+\aY});
\draw[-{Stealth[length=1.6mm,width=1.5mm]}, draw=accent, line width=0.8pt] (13.40,4.10) -- (13.40,{5.67+\aY});
\end{scope}

\begin{scope}[on background layer]
  \draw[rounded corners=4pt, draw=accent!60, fill=white, line width=0.6pt] (0.00,0.00) rectangle (17.40,4.10);
\end{scope}
\node[ptab, text=accent!90!black] at (0.30,4.10) {(b) Joint Region Judge Model};
\node[num, anchor=east] at (16.02,3.86) {\figice\ frozen};
\node[num, anchor=east] at (17.20,3.86) {\figfire\ trained};

\def\bsA{0.15}\def\bsB{0.30}\def\bsC{0.80}\def\bsD{0.85}\def\bsE{0.45}
\def\bsG{0.35}%
\def\bsW{0.55}\def\bsL{0.30}%
\begin{scope}[xshift=\bsA cm]
\node[pan] (crop) at (0.66,3.12) {\includegraphics[height=1.273cm]{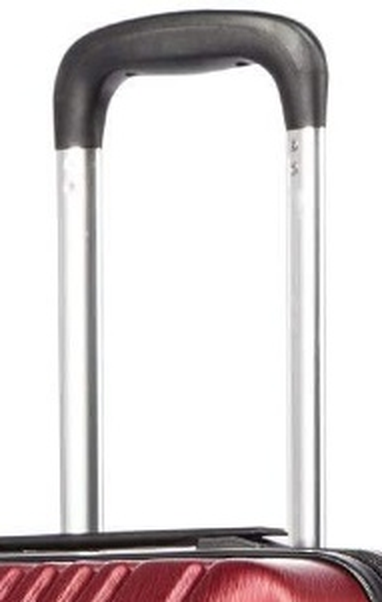}};
\node[pan] (regs) at (0.66,1.50) {\includegraphics[height=1.273cm]{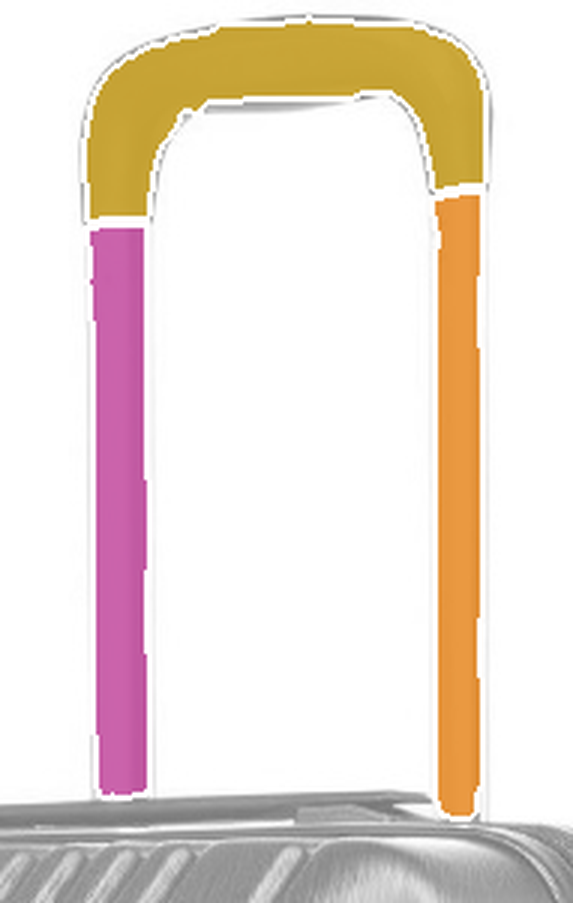}};
\node[num, anchor=north, align=center, font=\sffamily\fontsize{6.5}{7.2}\selectfont] at (regs.south) {candidate\\masks};
\end{scope}

\begin{scope}[xshift=\bsB cm]
\draw[lbox] (1.26,2.70) -- (2.42,2.90) -- (2.42,3.34) -- (1.26,3.54) -- cycle;
\node[text=lmcol, align=center, inner sep=0, font=\sffamily\fontsize{6.6}{7.4}\selectfont] at (1.73,3.12) {encoder};
\node[inner sep=0] at (2.29,3.12) {\figice};
\draw[lbox] (2.62,2.70) rectangle (3.78,3.54);
\node[text=lmcol, inner sep=0, font=\sffamily\fontsize{6.6}{7.4}\selectfont] at (3.20,3.12) {merger\,\figfire};
\draw[wire] (crop.east) -- (1.26,3.12);
\draw[wire] (2.42,3.12) -- (2.62,3.12);

\draw[abox, fill=accent!12, line width=0.7pt] (1.40,0.25) rectangle (4.05,2.50);
\node[pan, anchor=south west] (rtgrid) at (1.80,0.60) {\includegraphics[width=1.12cm]{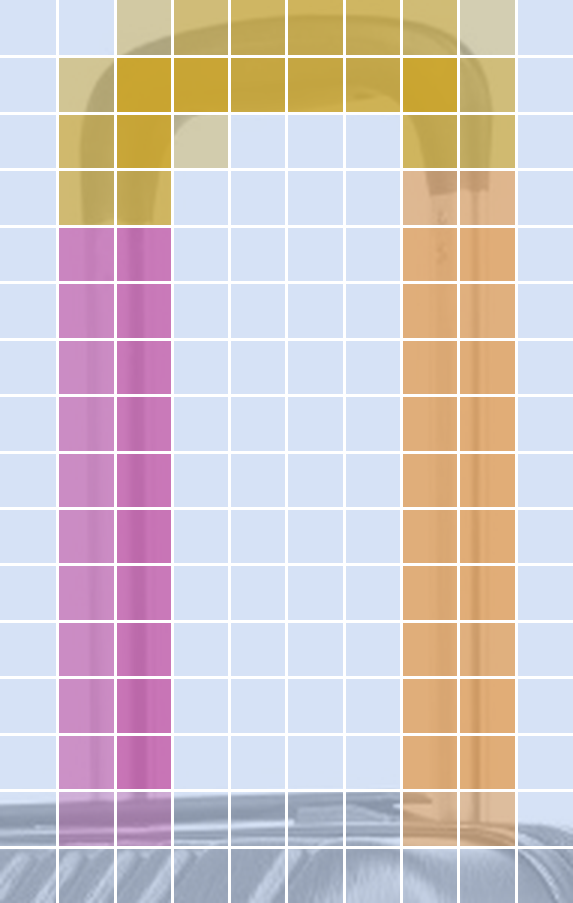}};
\node[mname, text=accent!90!black, anchor=south] at (2.725,0.28) {Region Tokenizer\,\figfire};
\draw[wire, draw=upcol!80] (2.80,2.70) -- (2.80,2.365);
\draw[wire] (regs.east |- 0,1.50) -- (1.80,1.50);
\foreach \y/\c in {2.30/rc1, 1.55/rc2, 0.80/rc6} {
  \draw[-{Stealth[length=1.0mm,width=0.9mm]}, draw=\c, line width=0.6pt] (2.94,\y) -- (3.51,\y);
  \node[circle, draw=\c, fill=white, line width=0.6pt, inner sep=0pt, minimum size=0.24cm, text=\c,
        font=\fontsize{6.5}{6.5}\selectfont] at (3.6425,\y) {$\Sigma$};
}
\end{scope}
\draw[wire, draw=upcol!80] ({3.78+\bsB},3.12) -- ({3.90+\bsC},3.12);

\begin{scope}[xshift=\bsC cm]
\node[inner sep=0, draw=upcol!85, line width=0.5pt] at (4.06,3.12) {\includegraphics[width=0.30cm]{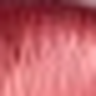}};
\node[inner sep=0, draw=upcol!85, line width=0.5pt] at (4.40,3.12) {\includegraphics[width=0.30cm]{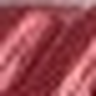}};
\node[inner sep=0, draw=upcol!85, line width=0.5pt] at (5.12,3.12) {\includegraphics[width=0.30cm]{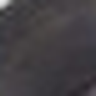}};
\node[num] at (4.76,3.12) {$\cdots$};
\node[rk, fill=rc1, anchor=west] (rt1) at (3.865,2.3) {$R_{1}$};
\node[inner sep=0, anchor=west] (ri1) at (rt1.east) {\includegraphics[height=0.34cm]{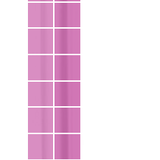}};
\draw[draw=rc1, line width=0.8pt, rounded corners=0.8pt] (rt1.north west) rectangle (ri1.south east);
\node[tchip, anchor=east, draw=rc1, line width=0.6pt] at (5.585,2.3) {Answer:};
\node[rk, fill=rc2, anchor=west] (rt2) at (3.865,1.55) {$R_{2}$};
\node[inner sep=0, anchor=west] (ri2) at (rt2.east) {\includegraphics[height=0.34cm]{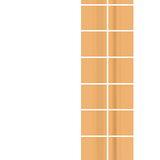}};
\draw[draw=rc2, line width=0.8pt, rounded corners=0.8pt] (rt2.north west) rectangle (ri2.south east);
\node[tchip, anchor=east, draw=rc2, line width=0.6pt] at (5.585,1.55) {Answer:};
\node[rk, fill=rc6, anchor=west] (rt3) at (3.865,0.8) {$R_{3}$};
\node[inner sep=0, anchor=west] (ri3) at (rt3.east) {\includegraphics[height=0.34cm]{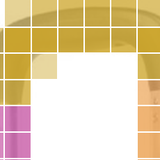}};
\draw[draw=rc6, line width=0.8pt, rounded corners=0.8pt] (rt3.north west) rectangle (ri3.south east);
\node[tchip, anchor=east, draw=rc6, line width=0.6pt] at (5.585,0.8) {Answer:};

\end{scope}
\foreach \y/\c in {2.3/rc1, 1.55/rc2, 0.8/rc6} {
  \draw[-{Stealth[length=1.0mm,width=0.9mm]}, draw=\c, line width=0.55pt] ({3.77+\bsB},\y) -- ({3.855+\bsC},\y);
  \draw[draw=\c!80, line width=0.45pt] ({5.585+\bsC},\y) -- ({5.85+\bsD},\y);
}
\draw[draw=upcol!70, line width=0.45pt] ({5.27+\bsC},3.12) -- ({5.85+\bsD},3.12);

\begin{scope}[xshift=\bsD cm]
\draw[lbox] (5.85,0.10) rectangle ({6.95+\bsL},3.82);
\node[font=\sffamily\fontsize{7.8}{8.8}\selectfont\bfseries, text=lmcol, rotate=90, inner sep=0] (llmt) at (6.33,1.79)
  {Large Language Model};
\node[inner sep=0, anchor=south] at ([yshift=1.2pt]llmt.east) {\figice};
\draw[abox, fill=accent!14] (6.80,0.24) rectangle (7.08,3.68);
\node[mname, text=accent!90!black, rotate=90, inner sep=0] (lorat) at (6.94,1.79) {LoRA};
\node[inner sep=0, anchor=south] at ([yshift=1.2pt]lorat.east) {\figfire};

\begin{scope}[xshift=\bsW cm]

\draw[rounded corners=3pt, draw=lmcol!40, fill=lmcol!5, line width=0.5pt] (7.085,0.40) rectangle (7.585,2.62);
\foreach \k/\y/\c in {1/2.3/rc1, 2/1.55/rc2, 3/0.8/rc6} {
  \draw[draw=\c, line width=0.45pt] ({6.95+\bsL-\bsW},\y) -- (7.155,\y);
  \node[rk, fill=\c, minimum width=0.36cm] at (7.335,\y) {$h_{\k}$};
}
\draw[-{Stealth[length=1.3mm,width=1.2mm]}, draw=inkgray!65, line width=0.6pt] (7.585,2.41) -- ({7.62+\bsG},2.41);
\draw[-{Stealth[length=1.3mm,width=1.2mm]}, draw=inkgray!65, line width=0.6pt] (7.585,0.575) -- ({7.62+\bsG},0.575);
\draw[-{Stealth[length=1.3mm,width=1.2mm]}, draw=upcol!80, line width=0.6pt] ({6.95+\bsL-\bsW},3.00) -- ({7.62+\bsG},3.00);
\node[text=upcol!85!black, anchor=south, inner sep=1pt, font=\sffamily\fontsize{6.6}{7.4}\selectfont]
  at ({(6.95+\bsL-\bsW+7.62+\bsG)/2},3.03) {features};

\begin{scope}[xshift=\bsG cm]
\draw[rounded corners=4pt, draw=accent!85, fill=accent!7, line width=0.8pt] (7.62,1.36) rectangle (10.80,3.46);
\node[font=\sffamily\fontsize{8}{9}\selectfont\bfseries, text=accent!90!black, anchor=west] at (7.72,3.28) {Joint decision head\,\figfire};
\node[pan, anchor=south west] (setimg) at (7.985,1.44) {\includegraphics[width=2.45cm]{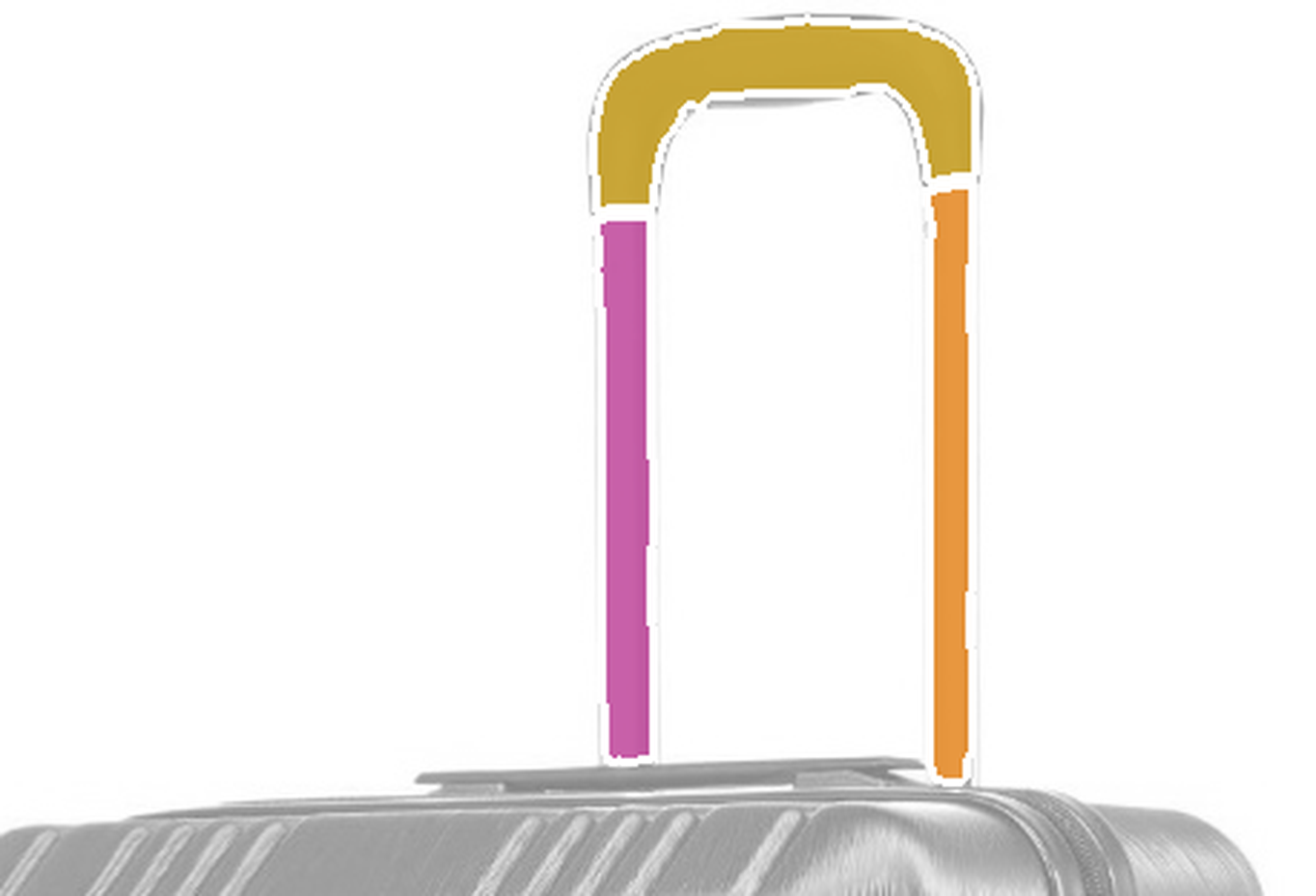}};
\coordinate (s1) at ($(7.985,1.44)+(0.4757*2.45,0.4574*1.667)$);
\coordinate (s2) at ($(7.985,1.44)+(0.7216*2.45,0.4637*1.667)$);
\coordinate (s3) at ($(7.985,1.44)+(0.5920*2.45,0.8978*1.667)$);
\foreach \a/\b/\bend in {s1/s2/22, s1/s3/-30, s3/s2/16} {
  \draw[white, line width=2.0pt, bend left=\bend, shorten <=1.6mm, shorten >=1.6mm] (\a) to (\b);
  \draw[{Stealth[length=1.3mm,width=1.2mm]}-{Stealth[length=1.3mm,width=1.2mm]}, draw=accent!90!black,
        line width=0.75pt, bend left=\bend, shorten <=1.6mm, shorten >=1.6mm] (\a) to (\b);
}
\foreach \s/\c in {s1/rc1, s2/rc2, s3/rc6}
  \node[tok, fill=\c, draw=white, line width=0.6pt, minimum width=0.20cm, minimum height=0.20cm] at (\s) {};

\begin{scope}[shift={(11.40,1.60)}]
  \fill[rejcol!6] (0,0) rectangle (1.72,1.72);
  \fill[acccol!16] (0.86,0.86) rectangle (1.72,1.72);
  \draw[draw=inkgray!50, line width=0.4pt, dashed, dash pattern=on 1.2pt off 1pt] (0.86,0) -- (0.86,1.72);
  \draw[draw=inkgray!50, line width=0.4pt, dashed, dash pattern=on 1.2pt off 1pt] (0,0.86) -- (1.72,0.86);
  \draw[draw=inkgray!70, line width=0.5pt] (0,1.72) -- (0,0) -- (1.72,0);
  \node[num, anchor=north] at (0.86,-0.02) {membership $r_k$};
  \node[num, anchor=south, rotate=90] at (-0.02,0.86) {purity $r'_k$};
  \node[font=\sffamily\bfseries\fontsize{9}{9}\selectfont, inner sep=0, text=acccol!80!black] at (1.06,1.53) {\checkmark};
  \foreach \m/\p/\c in {0.035/0.037/rc1, 0.018/0.023/rc2, 0.993/0.985/rc6}
    \node[circle, fill=\c, draw=white, line width=0.5pt, inner sep=0pt, minimum size=0.17cm] at (1.72*\m,1.72*\p) {};
\end{scope}
\node[lbadge] at (11.52,3.54) {2};
\node[lbadge] at (11.80,3.54) {3};
\draw[wire] (10.80,3.05) -- (11.36,3.05);
\end{scope}
\end{scope}
\end{scope}

\node[pan] (jout) at ({15.70+\bsE},2.45) {\includegraphics[height=1.85cm]{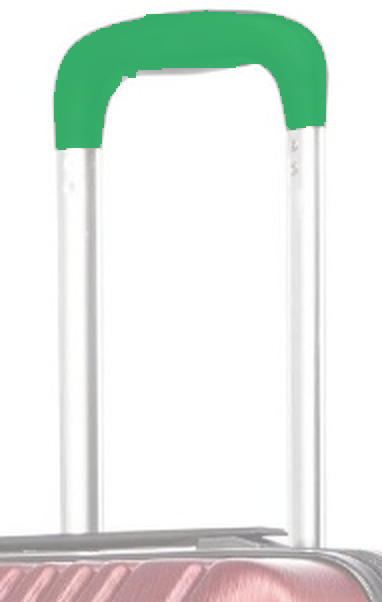}};
\node[num, anchor=south] at (jout.north) {$R_{3}$};
\draw[-{Stealth[length=1.6mm,width=1.4mm]}, draw=acccol, line width=0.9pt] ({13.30+\bsD+\bsW+\bsG},2.92) -- (jout.west |- 0,2.92);

\begin{scope}[xshift=\bsD cm]
\begin{scope}[xshift=\bsW cm]
\begin{scope}[xshift=\bsG cm]
\draw[rounded corners=4pt, draw=lmcol!85, fill=lmcol!7, line width=0.8pt] (7.62,0.10) rectangle (10.80,1.05);
\node[font=\sffamily\fontsize{8}{9}\selectfont\bfseries, text=lmcol, anchor=west, inner sep=0] at (7.72,0.87)
  {Verbalizer\,\figice};
\begin{scope}[shift={(8.45,0.44)}, xscale=0.95, yscale=0.58]
  \draw[draw=inkgray!40, line width=0.35pt, -{Stealth[length=0.8mm,width=0.7mm]}] (0,0) -- (0.62,0);
  \draw[draw=inkgray!40, line width=0.35pt, -{Stealth[length=0.8mm,width=0.7mm]}] (0,0) -- (0,0.46);
  \draw[draw=inkgray!40, line width=0.35pt, -{Stealth[length=0.8mm,width=0.7mm]}] (0,0) -- (-0.19,-0.27);
  \coordinate (eY) at (1.48,0.42);
  \coordinate (eN) at (1.48,-0.40);
  \shade[top color=acccol!85, bottom color=rejcol!85, middle color=gray!12] (1.445,-0.40) rectangle (1.515,0.42);
  \draw[-{Stealth[length=1.3mm,width=1.2mm]}, draw=acccol!85!black, line width=0.75pt, shorten >=0.2mm] (0,0) -- (eY);
  \draw[-{Stealth[length=1.3mm,width=1.2mm]}, draw=rejcol!85!black, line width=0.75pt, shorten >=0.2mm] (0,0) -- (eN);
  \node[num, text=acccol!70!black, anchor=west, inner sep=2pt] at (eY) {Yes};
  \node[num, text=rejcol!80!black, anchor=west, inner sep=2pt] at (eN) {No};
  \coordinate (hk) at (1.10,0.19);
  \coordinate (pk) at (1.48,0.19);
  \draw[-{Stealth[length=1.3mm,width=1.2mm]}, draw=lmcol, line width=0.85pt] (0,0) -- (hk);
  \draw[draw=lmcol!85, line width=0.5pt, dash pattern=on 1pt off 0.8pt] (hk) -- (pk);
  \fill[lmcol] (pk) circle (0.065);
  \node[num, text=lmcol, anchor=north, inner sep=1.2pt] at ($(hk)+(-0.12,-0.03)$) {$h_k$};
\end{scope}
\node[circle, draw=lmcol!85, fill=white, line width=0.6pt, inner sep=0pt, minimum size=0.27cm,
      font=\sffamily\bfseries\fontsize{7.4}{7.4}\selectfont, text=lmcol] (plus) at (11.085,0.575) {$+$};
\draw[-{Stealth[length=1.2mm,width=1.1mm]}, draw=lmcol!80, line width=0.6pt] (10.80,0.575) -- (plus.west);
\draw[-{Stealth[length=1.2mm,width=1.1mm]}, draw=accent!80, line width=0.6pt] (10.80,1.55) -- (11.085,1.55) -- (plus.north);
\node[num, anchor=west] at (11.115,1.12) {$r_k$};
\draw[rounded corners=2pt, draw=inkgray!55, fill=gray!8, line width=0.5pt] (11.37,0.405) rectangle (12.45,0.745);
\node[num] at (11.91,0.575) {softmax};
\draw[wire, draw=lmcol!80] (plus.east) -- (11.37,0.575);
\draw[wire, draw=lmcol!80] (12.45,0.575) -- (12.67,0.575);
\node[lbadge] at (12.82,0.575) {1};
\node[num, anchor=west] at (12.99,0.575) {object extent};
\node[pan, anchor=south west] at (14.45,0.17) {\includegraphics[width=0.50cm]{fig2cases/red_drag/h_object}};
\end{scope}
\end{scope}
\end{scope}
\end{tikzpicture}
\caption{\textbf{\method.} (a) A test query: one judge verifies each level's candidates at once (\checkmark/$\times$ per candidate).
No part passes, so the parts are split and sub-parts decide; an unsure object would backtrack (dashed).
(b) The judge (\cref{sec:judge}) on this sub-part request: in the joint decision head candidates attend to each other and must
pass both outputs; object choices also add the verbalizer, the projection of $h_k$ on the Yes$-$No embeddings.}
\label{fig:framework}
\end{figure*}
Given an RGB image $I$, an object name $o$ and an operation $a$, \method predicts a binary mask $\widehat Y$ on the
image grid; no box, point or example is given at test time. The method has two parts designed together. The
framework (\cref{sec:hierarchy}) turns a query into a short sequence of questions, each asking which regions of a
candidate set answer it. The judge (\cref{sec:judge}) is built for exactly these questions and trained on the requests
the framework itself produces (\cref{sec:training}). \Cref{fig:framework} shows both on one real query.

\subsection{Hierarchical verification framework}
\label{sec:hierarchy}
\smallhead{Requests and decision rules.}
Every request has the same form: a level $\ell$, an image or crop, candidate masks $\{M_k\}$ and the query
$(o,a)$, for which the judge returns one logit $z_k$ per candidate. At the selection levels, part and sub-part, a
candidate is accepted when $p_k=\sigma(z_k)>0.5$, and a foreground $F$ yields the union of what is accepted
together with the strongest verdict, which measures how well $F$ is supported:
\begin{equation}
 \widehat Y(F)=\bigcup_{k:\,p_k>0.5}M_k,\qquad e(F)=\max_k p_k .
 \label{eq:union}
\end{equation}
At the choice level, the object extent, exactly one candidate is right, and the probabilities are a softmax over
the logits of the group. Every judge we compare against is used through this interface and these rules. Because every decision
selects from a fixed candidate set, the output is always assembled from proposed regions: the segmenters supply its
boundaries and the judge supplies its meaning.

\smallhead{Object level.}
The query names the object, so object candidates come from SAM~3~\cite{carion2025sam3} prompted with $o$. When
instances reach the detector's own threshold of $0.5$, their union, with holes filled so that an object also covers
the region it encloses, such as a bowl's interior, is the object, and the object decision counts as settled.
Otherwise the three highest-scoring distinct masks (pairwise IoU at most $0.9$) form a choice question ---
\emph{which mask covers the complete object?} --- and the largest probability $\pi_{(1)}$ states how sure the object
decision is. Behind these hypotheses the ranking continues with class-agnostic regions from
PartGLEE~\cite{li2024partglee}, resolved into mutually exclusive pieces and visited only by backtracking. Settling the object first narrows every later question to a single object, so the part
questions compare only regions of the queried object.

\smallhead{Part and sub-part levels.}
For a foreground $F$ we crop $I$ around $F$ with $10\%$ padding and run PartGLEE with the generic text
\emph{part} on the crop with its exterior grayed out; the judge later sees the crop without the gray. Graying keeps PartGLEE from
proposing parts of neighbouring objects, while the ungrayed crop keeps the context the judge needs to recognise the
object and its use. Proposals are
clipped to $F$, given disjoint pixels and capped at six; discarded small proposals are merged into connected
clusters, and pieces of $F$ that no candidate covers become candidates of their own; SAM~3 masks of the part names
listed in the training data replace the generic parts they cover. The same call accepts any mask as foreground, so a part $M_i$
can be split once more into sub-parts clipped to $M_i$; since the crop shrinks with the region, a sub-part is seen at
a higher effective resolution than its parent.

\smallhead{Passing evidence across levels.}
\label{sec:passing}
The levels are not visited independently. Upward, a weak part verdict questions the object: if the object
decision is unsure ($\pi_{(1)}<\tau$) and its foreground is not confidently supported ($e(F_0)<\tau$), with
$\tau=0.9$, the search visits the next object hypotheses in order, up to $K=5$, and keeps
$F^\star=\arg\max_F e(F)$. Downward, every part of $F^\star$ is split and the sub-part verdicts refine the part
decision: if no part is accepted although the object is settled, the confirmed sub-parts become the prediction; if
a part is accepted, its rejected sub-parts are cut out of it, while a uniform verdict leaves it whole. The judge is
therefore asked three kinds of question --- choose the object extent, select parts, select the sub-parts of one
part --- and every answer is a subset of a candidate set. \Cref{alg:hv} summarises the procedure.

\begin{algorithm}[t]
\caption{Hierarchical verification}
\label{alg:hv}
\small
\begin{algorithmic}[1]
\Require image $I$, query $(o,a)$, judge $J$; $\tau=0.9$, $K=5$
\State $\mathcal{H}\gets$ object hypotheses of $o$: SAM~3 masks, then ranked PartGLEE regions
\If{the object is not settled} sort $\mathcal{H}$ by the extent choice $\pi=J(\mathcal{H})$ \EndIf
\State $F^\star\gets\mathcal{H}_1$;\quad $(\{M_k\},\{p_k\})\gets$ parts of $F^\star$ judged by $J$;\quad $e^\star\gets\max_k p_k$
\If{the object is not settled \textbf{and} $\pi_{(1)}<\tau$} \Comment{backtrack}
  \For{$F\in\mathcal{H}_2,\dots,\mathcal{H}_{K+1}$ \textbf{while} $e^\star<\tau$}
    \State judge the parts of $F$; \textbf{if} $e(F)>e^\star$ \textbf{then} $F^\star\gets F$ and update $\{M_k\},\{p_k\},e^\star$
  \EndFor
\EndIf
\State $A\gets\{k: p_k>0.5\}$ \Comment{sub-parts}
\If{$A=\emptyset$} \Comment{rescue}
  \State $\widehat Y\gets\bigcup_k\{S\in\mathrm{sub}(M_k): J(S)>0.5\}$
\Else \Comment{trim}
  \For{$k\in A$} remove from $M_k$ its rejected sub-parts, unless all or none are rejected \EndFor
  \State $\widehat Y\gets\bigcup_{k\in A}M_k$
\EndIf
\State \Return $\widehat Y$
\end{algorithmic}
\end{algorithm}

\subsection{Joint region judge}
\label{sec:judge}
These questions have three properties that general-purpose judges~\cite{valen2026,h2o2026lightning,simplejev2026jevany}
do not use. The answer depends on the pixels a candidate covers, often a small part of the image. It is a property
of the set, because the framework outputs the union of what it accepts. And the framework can supply about
twenty-five thousand training requests, too few to train a new scoring head from scratch. We build one judge around these
properties, starting from the raw Qwen3.5-4B weights~\cite{qwen2026}.

\smallhead{Request layout.}
The object name and operation come first, followed by a fixed sentence that names the level, such as \emph{The image
is a crop around the $o$; each candidate region is one part of the $o$}; only $o$ and $a$ are filled in. The image
follows, so that its visual tokens are encoded conditioned on the query, then a question asking \textsf{Yes} or
\textsf{No} for each region, and the candidates as rows \texttt{Region $k$:\,$R_k$\,Answer:}. A sub-part request
also carries the token of the parent part. Candidate $k$ is read at the final colon of its row.

\smallhead{Region tokenizer.}
Candidate $k$ enters the sequence as one token whose embedding is the coverage-weighted mean of the vision encoder's
own output over the token cells its mask touches, which already lives in the language model's input space, plus a
zero-initialised embedding $g$ of its box, area and centroid:
\begin{equation}
 R_k=\sum_{c} w_{kc}\,v_c+g(\text{geom}_k),\qquad w_{kc}\propto|M_k\cap c| .
 \label{eq:region}
\end{equation}
The judge therefore reads the pixels a candidate covers rather than inferring them from a drawn marker. Region
tokens sit at text positions after the image, so where a candidate lies and how large it is enter only through $g$. Because the token is pooled from features the language model
already reads, it needs no separate region encoder, and candidates of any shape or size are represented in the same
way.

\smallhead{Joint decision head.}
All candidates of a level share one sequence and one forward pass. This replaces one request with three images per
candidate by one request with one image per level: over the test queries the judge answers $6.3$ requests and
encodes $6.3$ images per query, against $23.8$ and $71.5$ for a per-candidate judge. Inside the causal language model a candidate sees only
those listed before it, and a joint decision head makes the comparison symmetric. For each candidate it takes the
readout state $h_k$ together with $p_k$, the last-layer image features pooled with the same weights $w_{kc}$, adds
one token for the final position of the sequence, and passes them through two bidirectional Transformer layers
without positional encoding. Two zero-initialised linear outputs give each candidate two logits, $r_k$ and $r'_k$,
used below; mixing is across the candidates of one request, never across levels. Seeing the whole level lets the head
lower the score of a fragment when a larger candidate already covers the queried part, and keep it when the fragment
is the only candidate on the part.

\smallhead{Verbalizer.}
When the judge chooses between object hypotheses, a verbalizer reads the decision from the model's own answer
tokens: candidate $k$ is scored by
\begin{equation}
 z_k=\alpha\,h_k^{\top}\!\left(E_{\textsf{Yes}}-E_{\textsf{No}}\right)+r_k ,
 \label{eq:readout}
\end{equation}
where $E_{\textsf{Yes}}$ and $E_{\textsf{No}}$ are the model's own tied output embeddings for the two answer tokens,
and the extent choice is a softmax over $z_k$, as in \cref{eq:union}. The first term is a scaled log-odds of
answering \textsf{Yes} rather than \textsf{No} at the readout position. It adds only the scale $\alpha$ and keeps the
pretrained notion of \textsf{Yes} available, instead of learning a scoring head from scratch.
At the part, sub-part and part-name levels the dataset's notion of a part differs from the language model's, so these decisions
come from the joint decision head alone, with $z_k=\min(r_k,r'_k)$ in \cref{eq:union}.

\subsection{Framework-consistent training}
\label{sec:training}
The judge must decide on the candidates the framework actually produces --- fragments, mixed regions and wrong
foregrounds --- and its probabilities must mean what the decision rule assumes. Training follows from these two
requirements.

\smallhead{Training requests.}
We run the frozen segmenters on the $1{,}230$ photos of the official InstructPart~\cite{wan2025instructpart}
training split and build requests from its $1{,}800$ training annotations only, at every level the framework asks
about: $1{,}158$ object-extent choices from SAM~3 prompted with the object name, labelled by the mask that best
matches the reviewed object and kept when that match reaches IoU $0.5$; $5{,}046$ part groups; and $14{,}730$
sub-part groups, resampled to $8{,}000$ per epoch. Two auxiliary tasks are never asked at test time and only shape
the shared representation: $2{,}307$ object-ownership choices among class-agnostic regions, and $2{,}112$ part-name
groups that ask which candidates form a named part, such as the handle of a fork. In total the judge is trained on
$25{,}353$ candidate groups.

\smallhead{Purity and membership targets.}
A candidate is accepted when its probability exceeds $0.5$, so that probability should estimate how much of the
candidate belongs to the answer. We therefore supervise $\sigma(r'_k)$ with the candidate's \emph{purity}, the
fraction of its reliably labelled pixels inside the queried part, through binary cross-entropy, and $\sigma(r_k)$
with a second binary cross-entropy on its membership in the IoU-optimal union over all subsets of the group; on
one shared logit, membership rather than purity would decide acceptance. Candidates with less than half of their pixels reliably labelled
are left out of the loss, and choice requests use cross-entropy on the softmax over the group. Purity tells the judge whether a candidate is
mostly target, and membership whether the best answer needs it; requiring both keeps small, clean fragments that
complete a part and rejects large regions that only touch it. Optimising the
expected IoU of the union directly, which seems to match the framework's output more closely, is worse
(supplementary material).

\smallhead{Optimisation.}
Starting from the raw weights, we train LoRA adapters~\cite{hu2022lora} in the language backbone together with the
visual merger, the geometry embedding and the joint decision head for two epochs, clipping gradients per parameter group
because the merger and geometry gradients are two orders of magnitude larger than the adapter ones. The judge's probabilities are used
as trained, without temperature scaling or any other calibration.

\section{Experiments}
\label{sec:experiments}

\subsection{Experimental settings}
\label{sec:setup}
\smallhead{Benchmarks.}
InstructPart~\cite{wan2025instructpart} is evaluated on its official test split: its $600$ test annotations become
$835$ object--operation queries over $44$ operations on $538$ photos, and the ground truth of a query is the union of
the object's native part masks labelled with that operation in either split. UMD~\cite{myers2015umd} contributes
$3{,}910$ queries on $1{,}922$ manually annotated frames covering all $17$ categories, sampled in proportion to each
object instance's frames and evenly in time, with void pixels ignored. The official IIT-AFF test
split~\cite{nguyen2017iit} contributes $3{,}680$ queries on $2{,}315$ images, restricted to queries whose target
pixels belong to a single object category. UMD and IIT-AFF are zero-shot for every method. The ablations of
\cref{sec:ablation} use the same test sets, training data and decision rules as the main tables. gIoU averages per-query
IoU, cIoU divides the summed intersection by the summed union, P$_{50}$ is the fraction of queries above IoU $0.5$,
and P$_{50\text{--}95}$ averages that fraction over thresholds $0.5$ to $0.95$.

\smallhead{Definition-aligned evaluation.}
The same operation word does not name the same region in every dataset. InstructPart marks a vessel's body for
\emph{grasp}, while UMD and IIT-AFF mark its handle; \emph{support} marks the load-bearing surface of furniture in
one and the blade of a shovel in the other. A judge that learns the training convention well is therefore
penalised on the transfer sets for doing so, which measures annotation style rather than grounding. All tables
therefore use \emph{aligned} words: on UMD and IIT-AFF, the operation text of the conflicting classes is replaced
through a mapping fixed before evaluation and derived from the native annotations alone, with no model scores or
metric search; \emph{grasp}, for instance, becomes \emph{hold}. The full mapping, and results under the original words on the earlier sets, are in the
supplementary material.

\smallhead{Baselines.}
Affordance-R1~\cite{wang2026affordancer1}, VisionReasoner~\cite{liu2026visionreasoner} and
AffordanceVLM~\cite{wu2025ragnet} use released weights and official inference paths.
OP-HRG~\cite{mehrab2026reasoning} uses released Qwen3-VL-4B weights, its native reasoning/self-check prompt and
SAM2-large. Its query is ``the part of the [object] used for this operation: [aligned operation]'': affordance
transfer without ground-truth part names or fine-tuning. We use greedy decoding capped at $2{,}048$ tokens
(BF16, SDPA). The released
System 1 judges Valen~\cite{valen2026}, H2O-Lightning~\cite{h2o2026lightning} and
JevAny-LoRA~\cite{simplejev2026jevany} are placed in our pipeline with the same candidates and rules. Every method
receives the same RGB, object name and operation. The System 1 judges therefore share candidates and decision rules
with \method, so the differences among those rows come from the judges alone.

\smallhead{Implementation details.}
The judge starts from the published Qwen3.5-4B weights, with LoRA rank $32$ ($\alpha=64$, no dropout) on every
linear layer of the language tower, a trainable visual merger, and the new geometry embedding and joint decision head; $101$M
parameters are updated. Learning rates are $10^{-4}$ for the adapters, $2\times10^{-5}$ for the merger and
$2\times10^{-4}$ for the new modules, AdamW with weight decay $0.01$, $100$ warm-up steps and a cosine schedule
floored at $2\%$. Each of two epochs sees all object, extent, part and part-name groups and a fresh sample of
$8{,}000$ sub-part groups, $18{,}623$ groups and $665$ updates per epoch; one group is one optimiser item, with
gradients accumulated over four on each of seven GPUs and clipped per parameter family. Images are capped between
$384^2$ and $640^2$ pixels, candidate order is shuffled, and crops are flipped with probability one half. Training
takes about one hour on seven RTX~4090 GPUs, and one trained judge with one set of hyperparameters serves all three
test sets. At test time the judge receives the candidates in the order the
segmenters produce them, without flipping or any other test-time augmentation.

\subsection{Main results}
\label{sec:main}
\Cref{tab:instructpart,tab:umd,tab:iit} compare \method with these methods under the aligned operation words, and
it leads every column of all three tables. Against the strongest other method on each set --- Affordance-R1 on
InstructPart, OP-HRG on IIT-AFF, and the released Valen on UMD --- the gIoU margins are $9.68$, $4.18$ and $7.26$
points, with paired 95\% bootstrap intervals of $[6.6, 12.8]$, $[3.2, 5.1]$ and $[5.0, 9.6]$ that resample photos or,
on UMD, object instances. In-domain cIoU reaches $79.24$, against $49.66$
for the strongest System 2 result, OP-HRG, showing a large gap under pixel-pooled scoring. The released Valen
remains the strongest baseline on UMD yet trails by $17.9$ gIoU in domain, motivating training on the framework's
own questions. The lead is not confined to a few easy queries: \method is also first on P$_{50}$ and
P$_{50\text{--}95}$ in every table, so it holds across IoU thresholds. The strongest baseline changes from set to set
--- Affordance-R1 in domain, the released Valen on UMD and OP-HRG on IIT-AFF --- while \method is first on all three,
so verifying candidate regions transfers more evenly than either reasoning in text or judging one candidate at a time.

\begin{figure*}[t]
\centering
\begin{tikzpicture}[x=1cm, y=1cm, font=\sffamily\fontsize{6}{7}\selectfont]
\def\fqs{1.585}\def\fqg{0.05}\def\fqp{1.635}%
\node[text=inkgray!80, font=\sffamily\fontsize{5.6}{6.4}\selectfont] at (7.475,0.50) {own inference paths};
\draw[inkgray!45, line width=0.4pt] (4.280,0.38) -- (10.670,0.38);
\node[text=inkgray!80, font=\sffamily\fontsize{5.6}{6.4}\selectfont] at (13.197,0.50) {System 1 judges in our pipeline};
\draw[inkgray!45, line width=0.4pt] (10.820,0.38) -- (15.575,0.38);
\node[text=inkgray, anchor=base] at (1.652,0.10) {RGB input};
\node[text=inkgray, anchor=base] at (3.288,0.10) {Ground truth};
\node[text=inkgray, anchor=base] at (5.022,0.10) {Affordance-R1};
\node[text=inkgray, anchor=base] at (6.657,0.10) {VisionReasoner};
\node[text=inkgray, anchor=base] at (8.292,0.10) {AffordanceVLM};
\node[text=inkgray, anchor=base] at (9.928,0.10) {OP-HRG};
\node[text=inkgray, anchor=base] at (11.562,0.10) {Valen};
\node[text=inkgray, anchor=base] at (13.197,0.10) {H2O-Lightning};
\node[text=inkgray, anchor=base] at (14.832,0.10) {JevAny-LoRA};
\node[text=cvprblue, font=\sffamily\bfseries\fontsize{6.2}{7}\selectfont, anchor=base] at (16.567,0.10) {\method\ (ours)};
\node[rotate=90, align=center, inner sep=0, font=\sffamily\fontsize{6.2}{7}\selectfont] at (0.60,-0.873) {\textbf{kettle}\\[-0.4pt]\emph{open}};
\node[inner sep=0, anchor=south west, draw=inkgray!25, line width=0.3pt] at (0.860,-1.665) {\includegraphics[width=\fqs cm]{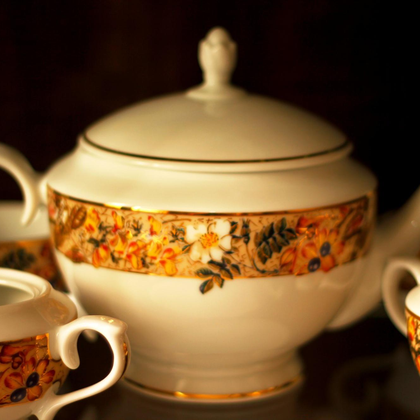}};
\node[inner sep=0, anchor=south west, draw=inkgray!25, line width=0.3pt] at (2.495,-1.665) {\includegraphics[width=\fqs cm]{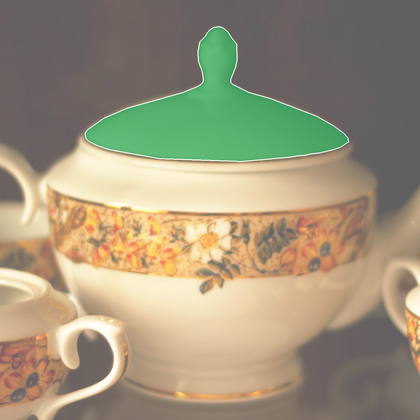}};
\node[inner sep=0, anchor=south west, draw=inkgray!25, line width=0.3pt] at (4.230,-1.665) {\includegraphics[width=\fqs cm]{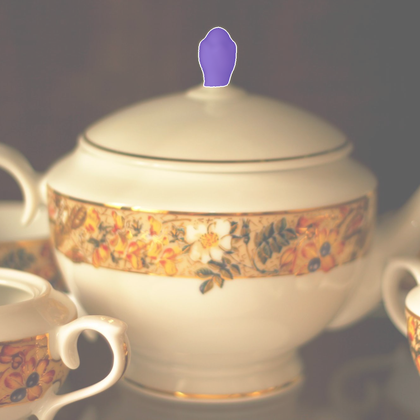}};
\node[anchor=south east, fill=white, fill opacity=0.85, text opacity=1, inner xsep=1.2pt, inner ysep=0.6pt, rounded corners=1pt, font=\sffamily\fontsize{5.6}{6}\selectfont] at (5.785,-1.635) {0.12};
\node[inner sep=0, anchor=south west, draw=inkgray!25, line width=0.3pt] at (5.865,-1.665) {\includegraphics[width=\fqs cm]{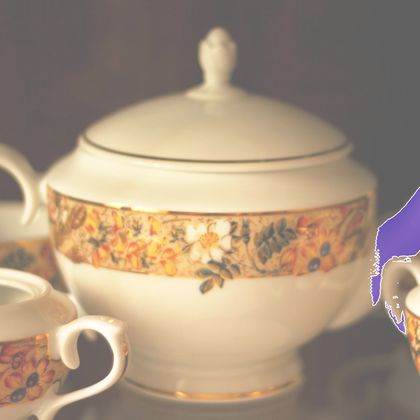}};
\node[anchor=south east, fill=white, fill opacity=0.85, text opacity=1, inner xsep=1.2pt, inner ysep=0.6pt, rounded corners=1pt, font=\sffamily\fontsize{5.6}{6}\selectfont] at (7.420,-1.635) {0.00};
\node[inner sep=0, anchor=south west, draw=inkgray!25, line width=0.3pt] at (7.500,-1.665) {\includegraphics[width=\fqs cm]{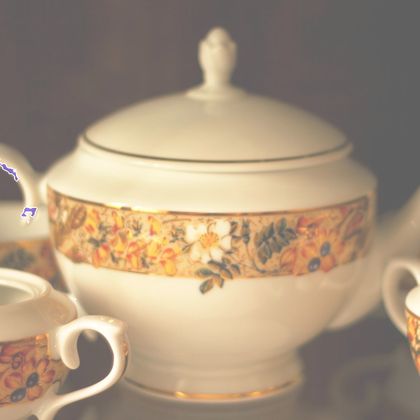}};
\node[anchor=south east, fill=white, fill opacity=0.85, text opacity=1, inner xsep=1.2pt, inner ysep=0.6pt, rounded corners=1pt, font=\sffamily\fontsize{5.6}{6}\selectfont] at (9.055,-1.635) {0.00};
\node[inner sep=0, anchor=south west, draw=inkgray!25, line width=0.3pt] at (9.135,-1.665) {\includegraphics[width=\fqs cm]{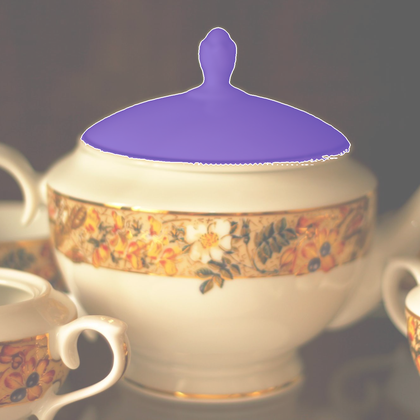}};
\node[anchor=south east, fill=white, fill opacity=0.85, text opacity=1, inner xsep=1.2pt, inner ysep=0.6pt, rounded corners=1pt, font=\sffamily\fontsize{5.6}{6}\selectfont] at (10.690,-1.635) {0.91};
\node[inner sep=0, anchor=south west, draw=inkgray!25, line width=0.3pt] at (10.770,-1.665) {\includegraphics[width=\fqs cm]{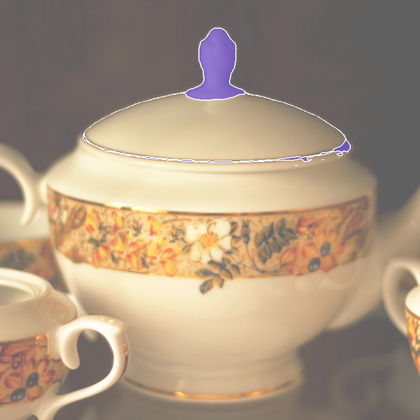}};
\node[anchor=south east, fill=white, fill opacity=0.85, text opacity=1, inner xsep=1.2pt, inner ysep=0.6pt, rounded corners=1pt, font=\sffamily\fontsize{5.6}{6}\selectfont] at (12.325,-1.635) {0.16};
\node[inner sep=0, anchor=south west, draw=inkgray!25, line width=0.3pt] at (12.405,-1.665) {\includegraphics[width=\fqs cm]{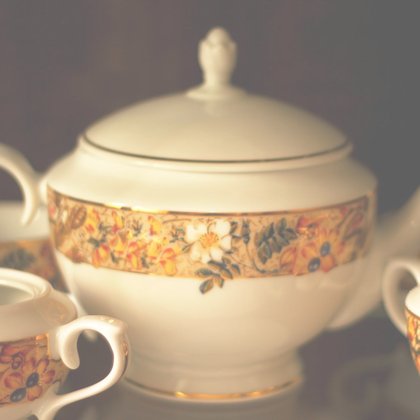}};
\node[anchor=south east, fill=white, fill opacity=0.85, text opacity=1, inner xsep=1.2pt, inner ysep=0.6pt, rounded corners=1pt, font=\sffamily\fontsize{5.6}{6}\selectfont] at (13.960,-1.635) {0.00};
\node[inner sep=0, anchor=south west, draw=inkgray!25, line width=0.3pt] at (14.040,-1.665) {\includegraphics[width=\fqs cm]{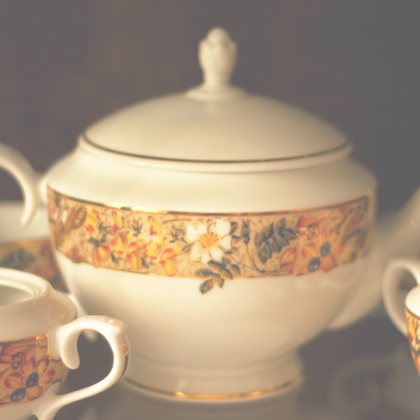}};
\node[anchor=south east, fill=white, fill opacity=0.85, text opacity=1, inner xsep=1.2pt, inner ysep=0.6pt, rounded corners=1pt, font=\sffamily\fontsize{5.6}{6}\selectfont] at (15.595,-1.635) {0.00};
\node[inner sep=0, anchor=south west, draw=cvprblue, line width=0.9pt] at (15.775,-1.665) {\includegraphics[width=\fqs cm]{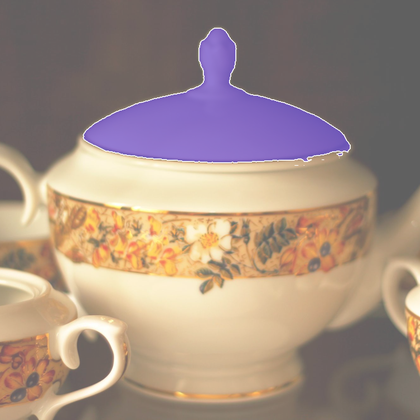}};
\node[anchor=south east, fill=white, fill opacity=0.85, text opacity=1, inner xsep=1.2pt, inner ysep=0.6pt, rounded corners=1pt, font=\sffamily\bfseries\fontsize{5.6}{6}\selectfont] at (17.330,-1.635) {0.94};
\node[rotate=90, align=center, inner sep=0, font=\sffamily\fontsize{6.2}{7}\selectfont] at (0.60,-2.507) {\textbf{wine glass}\\[-0.4pt]\emph{hold}};
\node[inner sep=0, anchor=south west, draw=inkgray!25, line width=0.3pt] at (0.860,-3.300) {\includegraphics[width=\fqs cm]{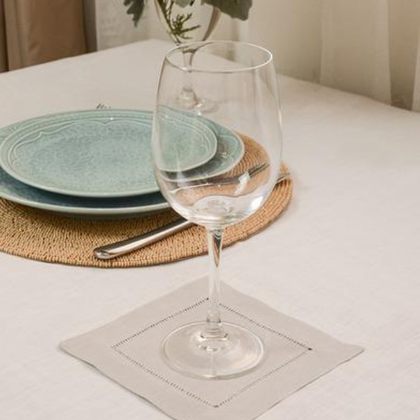}};
\node[inner sep=0, anchor=south west, draw=inkgray!25, line width=0.3pt] at (2.495,-3.300) {\includegraphics[width=\fqs cm]{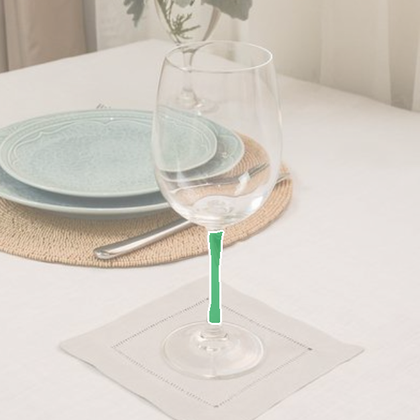}};
\node[inner sep=0, anchor=south west, draw=inkgray!25, line width=0.3pt] at (4.230,-3.300) {\includegraphics[width=\fqs cm]{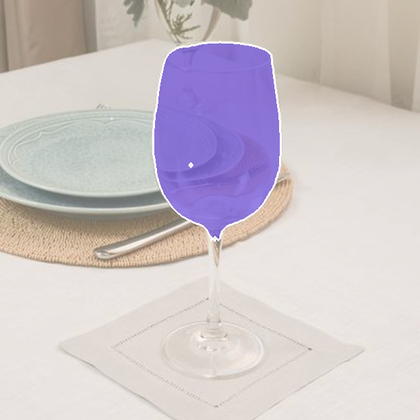}};
\node[anchor=south east, fill=white, fill opacity=0.85, text opacity=1, inner xsep=1.2pt, inner ysep=0.6pt, rounded corners=1pt, font=\sffamily\fontsize{5.6}{6}\selectfont] at (5.785,-3.270) {0.01};
\node[inner sep=0, anchor=south west, draw=inkgray!25, line width=0.3pt] at (5.865,-3.300) {\includegraphics[width=\fqs cm]{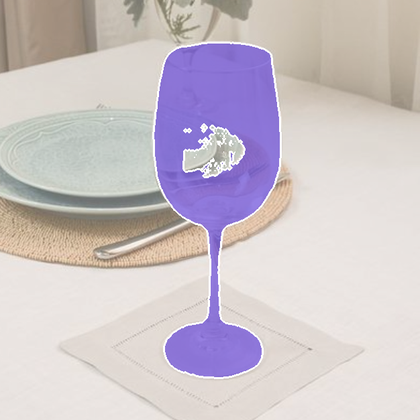}};
\node[anchor=south east, fill=white, fill opacity=0.85, text opacity=1, inner xsep=1.2pt, inner ysep=0.6pt, rounded corners=1pt, font=\sffamily\fontsize{5.6}{6}\selectfont] at (7.420,-3.270) {0.05};
\node[inner sep=0, anchor=south west, draw=inkgray!25, line width=0.3pt] at (7.500,-3.300) {\includegraphics[width=\fqs cm]{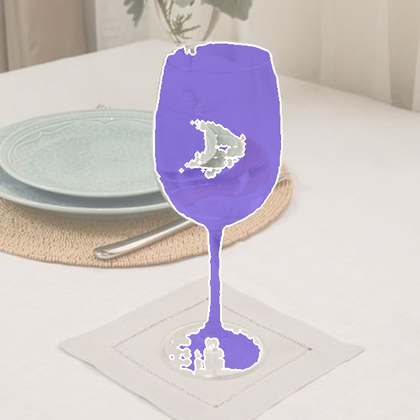}};
\node[anchor=south east, fill=white, fill opacity=0.85, text opacity=1, inner xsep=1.2pt, inner ysep=0.6pt, rounded corners=1pt, font=\sffamily\fontsize{5.6}{6}\selectfont] at (9.055,-3.270) {0.05};
\node[inner sep=0, anchor=south west, draw=inkgray!25, line width=0.3pt] at (9.135,-3.300) {\includegraphics[width=\fqs cm]{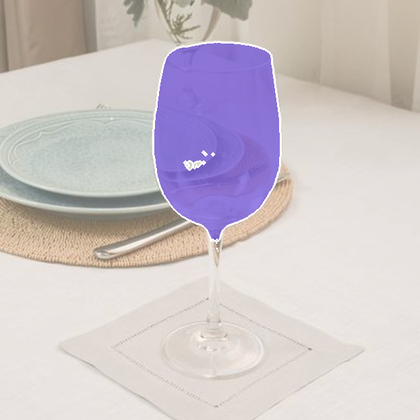}};
\node[anchor=south east, fill=white, fill opacity=0.85, text opacity=1, inner xsep=1.2pt, inner ysep=0.6pt, rounded corners=1pt, font=\sffamily\fontsize{5.6}{6}\selectfont] at (10.690,-3.270) {0.01};
\node[inner sep=0, anchor=south west, draw=inkgray!25, line width=0.3pt] at (10.770,-3.300) {\includegraphics[width=\fqs cm]{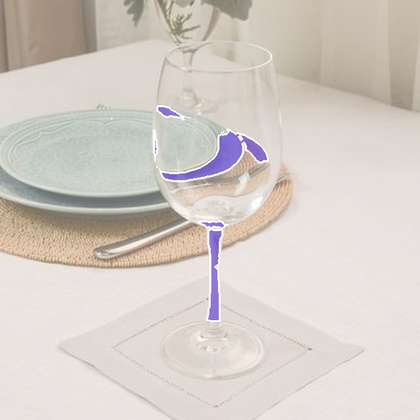}};
\node[anchor=south east, fill=white, fill opacity=0.85, text opacity=1, inner xsep=1.2pt, inner ysep=0.6pt, rounded corners=1pt, font=\sffamily\fontsize{5.6}{6}\selectfont] at (12.325,-3.270) {0.28};
\node[inner sep=0, anchor=south west, draw=inkgray!25, line width=0.3pt] at (12.405,-3.300) {\includegraphics[width=\fqs cm]{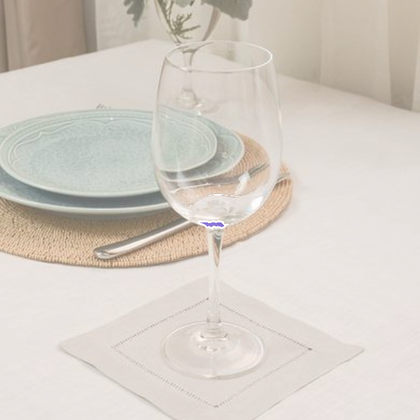}};
\node[anchor=south east, fill=white, fill opacity=0.85, text opacity=1, inner xsep=1.2pt, inner ysep=0.6pt, rounded corners=1pt, font=\sffamily\fontsize{5.6}{6}\selectfont] at (13.960,-3.270) {0.00};
\node[inner sep=0, anchor=south west, draw=inkgray!25, line width=0.3pt] at (14.040,-3.300) {\includegraphics[width=\fqs cm]{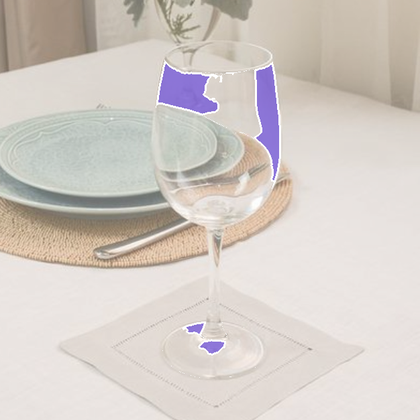}};
\node[anchor=south east, fill=white, fill opacity=0.85, text opacity=1, inner xsep=1.2pt, inner ysep=0.6pt, rounded corners=1pt, font=\sffamily\fontsize{5.6}{6}\selectfont] at (15.595,-3.270) {0.00};
\node[inner sep=0, anchor=south west, draw=cvprblue, line width=0.9pt] at (15.775,-3.300) {\includegraphics[width=\fqs cm]{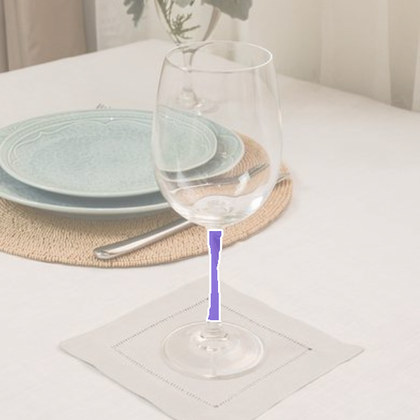}};
\node[anchor=south east, fill=white, fill opacity=0.85, text opacity=1, inner xsep=1.2pt, inner ysep=0.6pt, rounded corners=1pt, font=\sffamily\bfseries\fontsize{5.6}{6}\selectfont] at (17.330,-3.270) {0.86};
\node[rotate=90, align=center, inner sep=0, font=\sffamily\fontsize{6.2}{7}\selectfont] at (0.60,-4.143) {\textbf{oven}\\[-0.4pt]\emph{open}};
\node[inner sep=0, anchor=south west, draw=inkgray!25, line width=0.3pt] at (0.860,-4.935) {\includegraphics[width=\fqs cm]{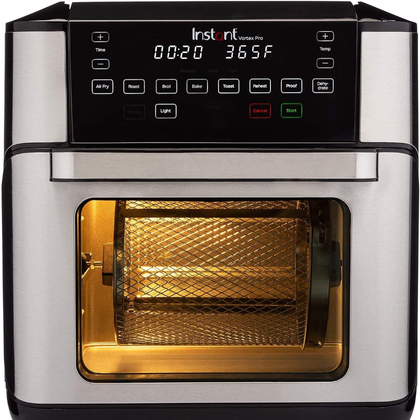}};
\node[inner sep=0, anchor=south west, draw=inkgray!25, line width=0.3pt] at (2.495,-4.935) {\includegraphics[width=\fqs cm]{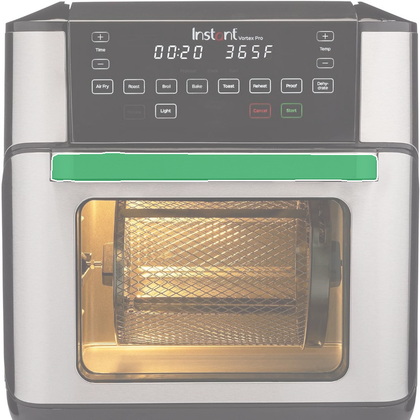}};
\node[inner sep=0, anchor=south west, draw=inkgray!25, line width=0.3pt] at (4.230,-4.935) {\includegraphics[width=\fqs cm]{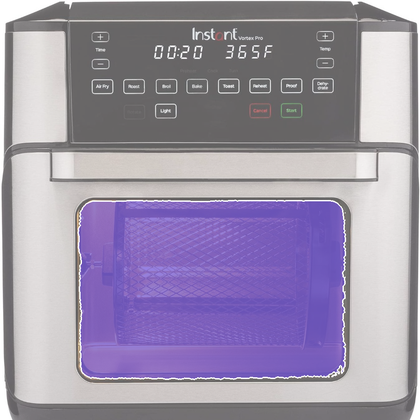}};
\node[anchor=south east, fill=white, fill opacity=0.85, text opacity=1, inner xsep=1.2pt, inner ysep=0.6pt, rounded corners=1pt, font=\sffamily\fontsize{5.6}{6}\selectfont] at (5.785,-4.905) {0.00};
\node[inner sep=0, anchor=south west, draw=inkgray!25, line width=0.3pt] at (5.865,-4.935) {\includegraphics[width=\fqs cm]{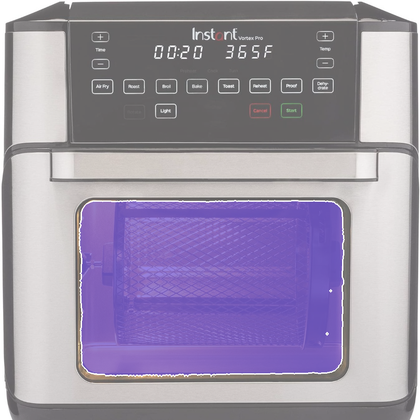}};
\node[anchor=south east, fill=white, fill opacity=0.85, text opacity=1, inner xsep=1.2pt, inner ysep=0.6pt, rounded corners=1pt, font=\sffamily\fontsize{5.6}{6}\selectfont] at (7.420,-4.905) {0.00};
\node[inner sep=0, anchor=south west, draw=inkgray!25, line width=0.3pt] at (7.500,-4.935) {\includegraphics[width=\fqs cm]{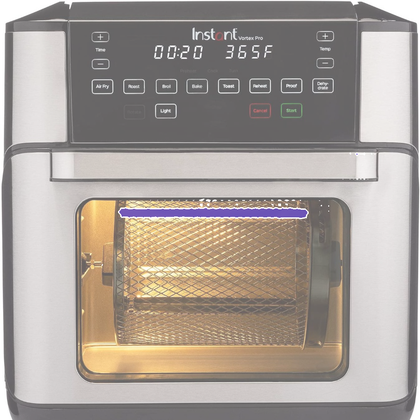}};
\node[anchor=south east, fill=white, fill opacity=0.85, text opacity=1, inner xsep=1.2pt, inner ysep=0.6pt, rounded corners=1pt, font=\sffamily\fontsize{5.6}{6}\selectfont] at (9.055,-4.905) {0.00};
\node[inner sep=0, anchor=south west, draw=inkgray!25, line width=0.3pt] at (9.135,-4.935) {\includegraphics[width=\fqs cm]{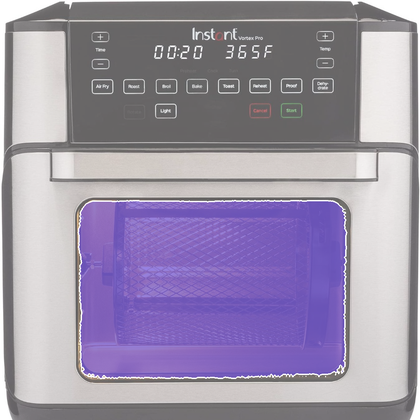}};
\node[anchor=south east, fill=white, fill opacity=0.85, text opacity=1, inner xsep=1.2pt, inner ysep=0.6pt, rounded corners=1pt, font=\sffamily\fontsize{5.6}{6}\selectfont] at (10.690,-4.905) {0.00};
\node[inner sep=0, anchor=south west, draw=inkgray!25, line width=0.3pt] at (10.770,-4.935) {\includegraphics[width=\fqs cm]{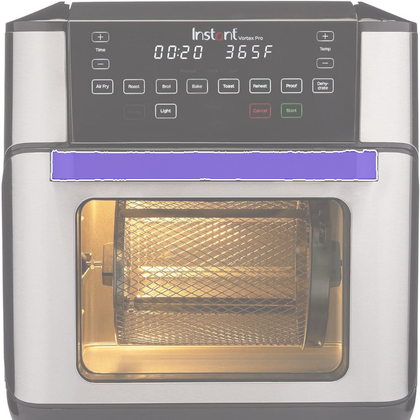}};
\node[anchor=south east, fill=white, fill opacity=0.85, text opacity=1, inner xsep=1.2pt, inner ysep=0.6pt, rounded corners=1pt, font=\sffamily\fontsize{5.6}{6}\selectfont] at (12.325,-4.905) {0.91};
\node[inner sep=0, anchor=south west, draw=inkgray!25, line width=0.3pt] at (12.405,-4.935) {\includegraphics[width=\fqs cm]{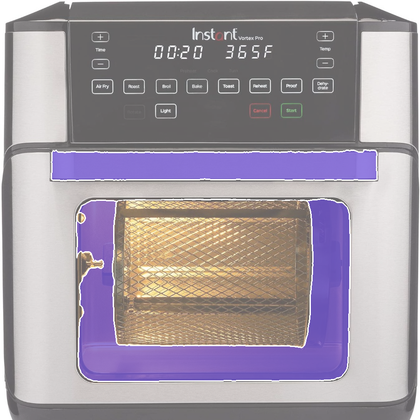}};
\node[anchor=south east, fill=white, fill opacity=0.85, text opacity=1, inner xsep=1.2pt, inner ysep=0.6pt, rounded corners=1pt, font=\sffamily\fontsize{5.6}{6}\selectfont] at (13.960,-4.905) {0.29};
\node[inner sep=0, anchor=south west, draw=inkgray!25, line width=0.3pt] at (14.040,-4.935) {\includegraphics[width=\fqs cm]{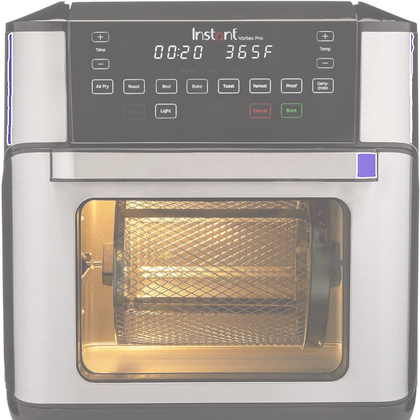}};
\node[anchor=south east, fill=white, fill opacity=0.85, text opacity=1, inner xsep=1.2pt, inner ysep=0.6pt, rounded corners=1pt, font=\sffamily\fontsize{5.6}{6}\selectfont] at (15.595,-4.905) {0.05};
\node[inner sep=0, anchor=south west, draw=cvprblue, line width=0.9pt] at (15.775,-4.935) {\includegraphics[width=\fqs cm]{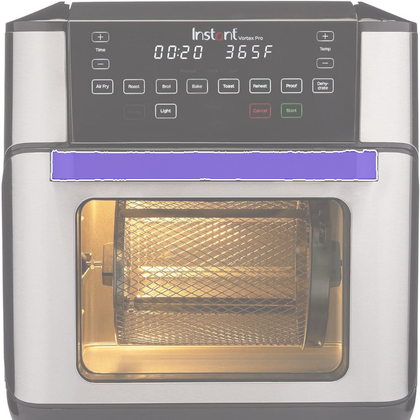}};
\node[anchor=south east, fill=white, fill opacity=0.85, text opacity=1, inner xsep=1.2pt, inner ysep=0.6pt, rounded corners=1pt, font=\sffamily\bfseries\fontsize{5.6}{6}\selectfont] at (17.330,-4.905) {0.91};
\node[rotate=90, align=center, inner sep=0, font=\sffamily\fontsize{6.2}{7}\selectfont] at (0.60,-5.777) {\textbf{mug}\\[-0.4pt]\emph{grasp}};
\node[inner sep=0, anchor=south west, draw=inkgray!25, line width=0.3pt] at (0.860,-6.570) {\includegraphics[width=\fqs cm]{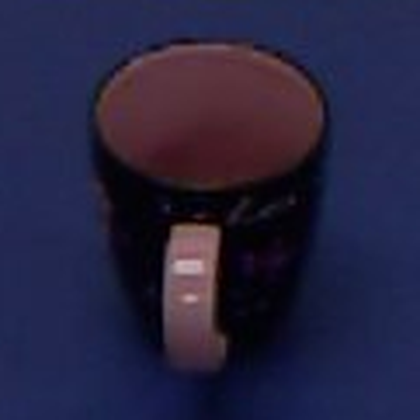}};
\node[inner sep=0, anchor=south west, draw=inkgray!25, line width=0.3pt] at (2.495,-6.570) {\includegraphics[width=\fqs cm]{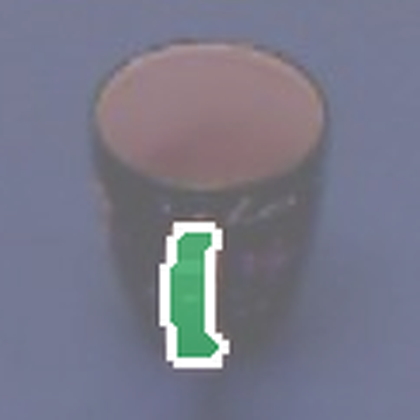}};
\node[inner sep=0, anchor=south west, draw=inkgray!25, line width=0.3pt] at (4.230,-6.570) {\includegraphics[width=\fqs cm]{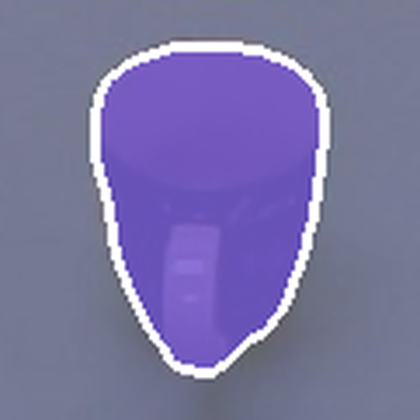}};
\node[anchor=south east, fill=white, fill opacity=0.85, text opacity=1, inner xsep=1.2pt, inner ysep=0.6pt, rounded corners=1pt, font=\sffamily\fontsize{5.6}{6}\selectfont] at (5.785,-6.540) {0.13};
\node[inner sep=0, anchor=south west, draw=inkgray!25, line width=0.3pt] at (5.865,-6.570) {\includegraphics[width=\fqs cm]{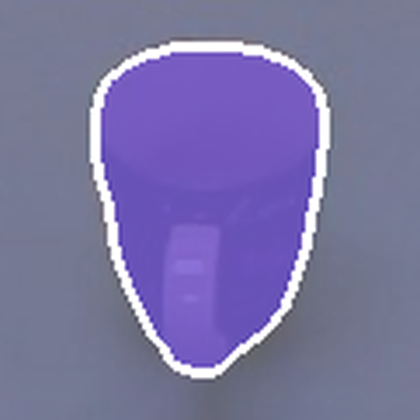}};
\node[anchor=south east, fill=white, fill opacity=0.85, text opacity=1, inner xsep=1.2pt, inner ysep=0.6pt, rounded corners=1pt, font=\sffamily\fontsize{5.6}{6}\selectfont] at (7.420,-6.540) {0.13};
\node[inner sep=0, anchor=south west, draw=inkgray!25, line width=0.3pt] at (7.500,-6.570) {\includegraphics[width=\fqs cm]{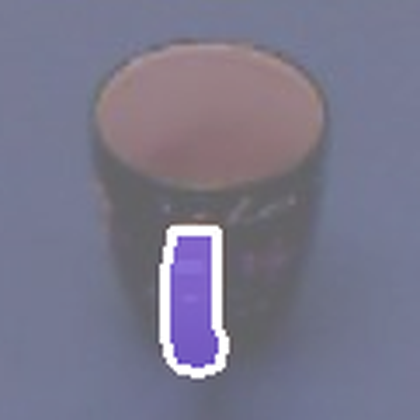}};
\node[anchor=south east, fill=white, fill opacity=0.85, text opacity=1, inner xsep=1.2pt, inner ysep=0.6pt, rounded corners=1pt, font=\sffamily\fontsize{5.6}{6}\selectfont] at (9.055,-6.540) {0.84};
\node[inner sep=0, anchor=south west, draw=inkgray!25, line width=0.3pt] at (9.135,-6.570) {\includegraphics[width=\fqs cm]{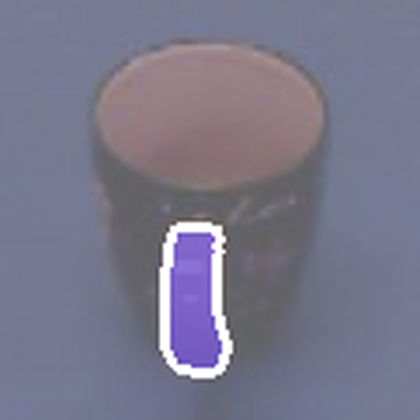}};
\node[anchor=south east, fill=white, fill opacity=0.85, text opacity=1, inner xsep=1.2pt, inner ysep=0.6pt, rounded corners=1pt, font=\sffamily\fontsize{5.6}{6}\selectfont] at (10.690,-6.540) {0.82};
\node[inner sep=0, anchor=south west, draw=inkgray!25, line width=0.3pt] at (10.770,-6.570) {\includegraphics[width=\fqs cm]{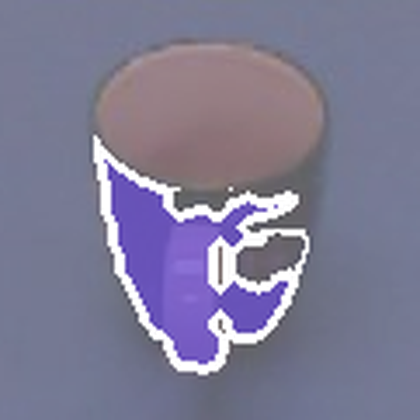}};
\node[anchor=south east, fill=white, fill opacity=0.85, text opacity=1, inner xsep=1.2pt, inner ysep=0.6pt, rounded corners=1pt, font=\sffamily\fontsize{5.6}{6}\selectfont] at (12.325,-6.540) {0.31};
\node[inner sep=0, anchor=south west, draw=inkgray!25, line width=0.3pt] at (12.405,-6.570) {\includegraphics[width=\fqs cm]{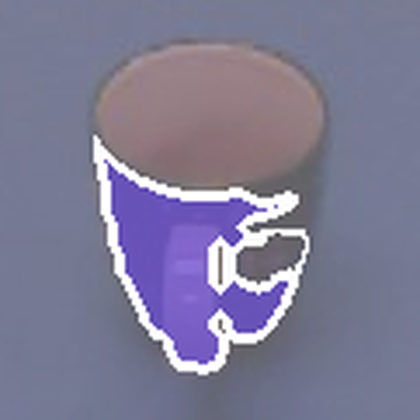}};
\node[anchor=south east, fill=white, fill opacity=0.85, text opacity=1, inner xsep=1.2pt, inner ysep=0.6pt, rounded corners=1pt, font=\sffamily\fontsize{5.6}{6}\selectfont] at (13.960,-6.540) {0.29};
\node[inner sep=0, anchor=south west, draw=inkgray!25, line width=0.3pt] at (14.040,-6.570) {\includegraphics[width=\fqs cm]{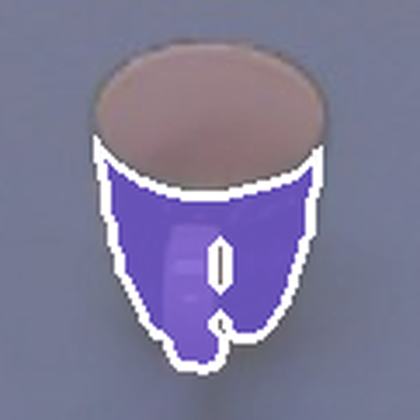}};
\node[anchor=south east, fill=white, fill opacity=0.85, text opacity=1, inner xsep=1.2pt, inner ysep=0.6pt, rounded corners=1pt, font=\sffamily\fontsize{5.6}{6}\selectfont] at (15.595,-6.540) {0.24};
\node[inner sep=0, anchor=south west, draw=cvprblue, line width=0.9pt] at (15.775,-6.570) {\includegraphics[width=\fqs cm]{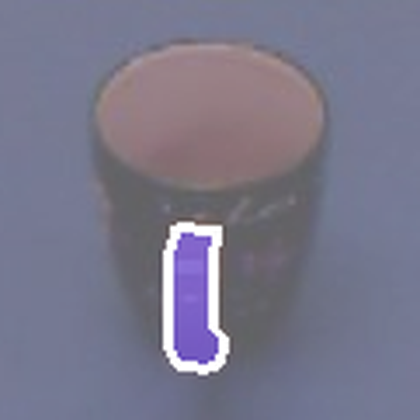}};
\node[anchor=south east, fill=white, fill opacity=0.85, text opacity=1, inner xsep=1.2pt, inner ysep=0.6pt, rounded corners=1pt, font=\sffamily\bfseries\fontsize{5.6}{6}\selectfont] at (17.330,-6.540) {0.87};
\node[rotate=90, align=center, inner sep=0, font=\sffamily\fontsize{6.2}{7}\selectfont] at (0.60,-7.412) {\textbf{pan}\\[-0.4pt]\emph{contain}};
\node[inner sep=0, anchor=south west, draw=inkgray!25, line width=0.3pt] at (0.860,-8.205) {\includegraphics[width=\fqs cm]{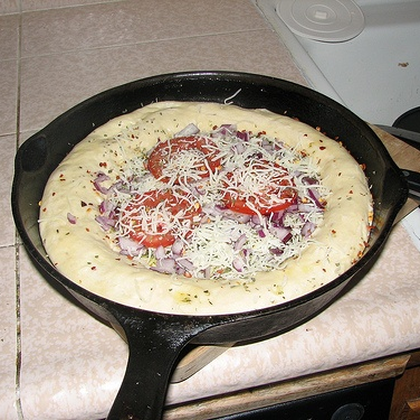}};
\node[inner sep=0, anchor=south west, draw=inkgray!25, line width=0.3pt] at (2.495,-8.205) {\includegraphics[width=\fqs cm]{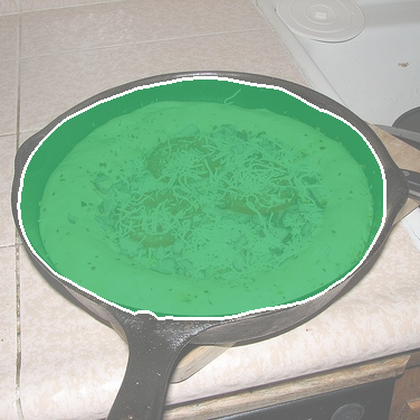}};
\node[inner sep=0, anchor=south west, draw=inkgray!25, line width=0.3pt] at (4.230,-8.205) {\includegraphics[width=\fqs cm]{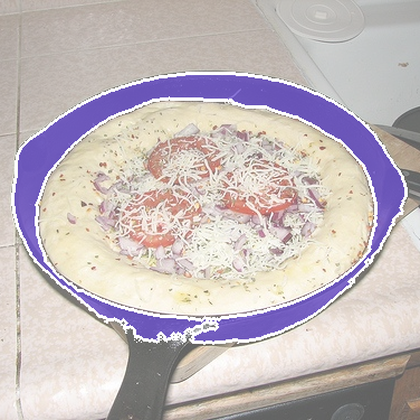}};
\node[anchor=south east, fill=white, fill opacity=0.85, text opacity=1, inner xsep=1.2pt, inner ysep=0.6pt, rounded corners=1pt, font=\sffamily\fontsize{5.6}{6}\selectfont] at (5.785,-8.175) {0.16};
\node[inner sep=0, anchor=south west, draw=inkgray!25, line width=0.3pt] at (5.865,-8.205) {\includegraphics[width=\fqs cm]{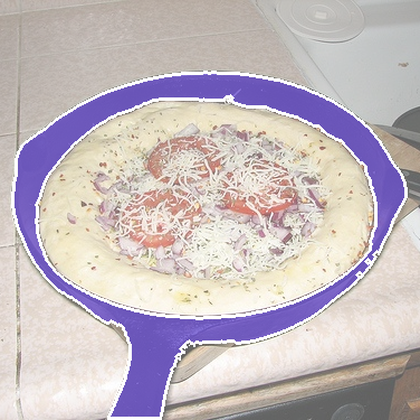}};
\node[anchor=south east, fill=white, fill opacity=0.85, text opacity=1, inner xsep=1.2pt, inner ysep=0.6pt, rounded corners=1pt, font=\sffamily\fontsize{5.6}{6}\selectfont] at (7.420,-8.175) {0.15};
\node[inner sep=0, anchor=south west, draw=inkgray!25, line width=0.3pt] at (7.500,-8.205) {\includegraphics[width=\fqs cm]{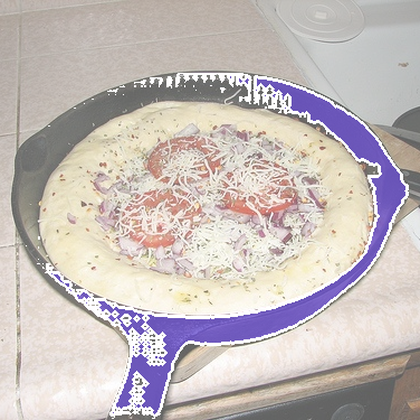}};
\node[anchor=south east, fill=white, fill opacity=0.85, text opacity=1, inner xsep=1.2pt, inner ysep=0.6pt, rounded corners=1pt, font=\sffamily\fontsize{5.6}{6}\selectfont] at (9.055,-8.175) {0.07};
\node[inner sep=0, anchor=south west, draw=inkgray!25, line width=0.3pt] at (9.135,-8.205) {\includegraphics[width=\fqs cm]{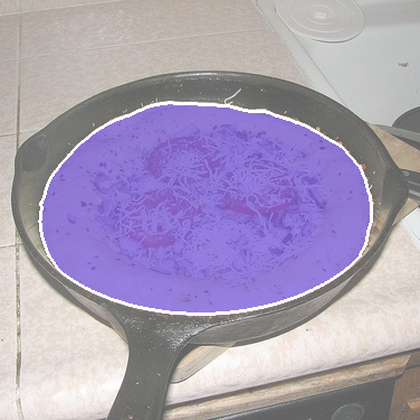}};
\node[anchor=south east, fill=white, fill opacity=0.85, text opacity=1, inner xsep=1.2pt, inner ysep=0.6pt, rounded corners=1pt, font=\sffamily\fontsize{5.6}{6}\selectfont] at (10.690,-8.175) {0.80};
\node[inner sep=0, anchor=south west, draw=inkgray!25, line width=0.3pt] at (10.770,-8.205) {\includegraphics[width=\fqs cm]{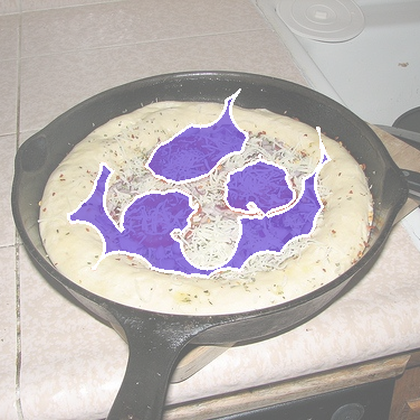}};
\node[anchor=south east, fill=white, fill opacity=0.85, text opacity=1, inner xsep=1.2pt, inner ysep=0.6pt, rounded corners=1pt, font=\sffamily\fontsize{5.6}{6}\selectfont] at (12.325,-8.175) {0.27};
\node[inner sep=0, anchor=south west, draw=inkgray!25, line width=0.3pt] at (12.405,-8.205) {\includegraphics[width=\fqs cm]{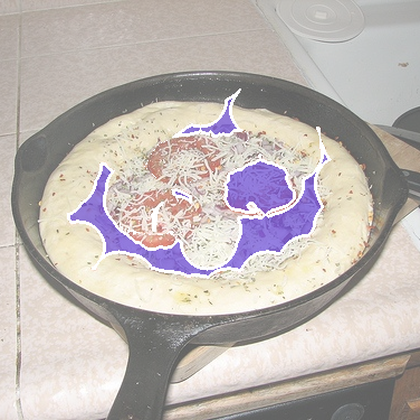}};
\node[anchor=south east, fill=white, fill opacity=0.85, text opacity=1, inner xsep=1.2pt, inner ysep=0.6pt, rounded corners=1pt, font=\sffamily\fontsize{5.6}{6}\selectfont] at (13.960,-8.175) {0.18};
\node[inner sep=0, anchor=south west, draw=inkgray!25, line width=0.3pt] at (14.040,-8.205) {\includegraphics[width=\fqs cm]{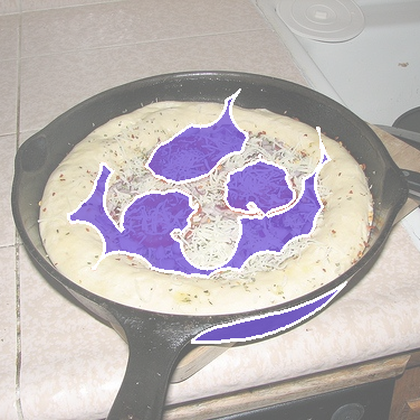}};
\node[anchor=south east, fill=white, fill opacity=0.85, text opacity=1, inner xsep=1.2pt, inner ysep=0.6pt, rounded corners=1pt, font=\sffamily\fontsize{5.6}{6}\selectfont] at (15.595,-8.175) {0.26};
\node[inner sep=0, anchor=south west, draw=cvprblue, line width=0.9pt] at (15.775,-8.205) {\includegraphics[width=\fqs cm]{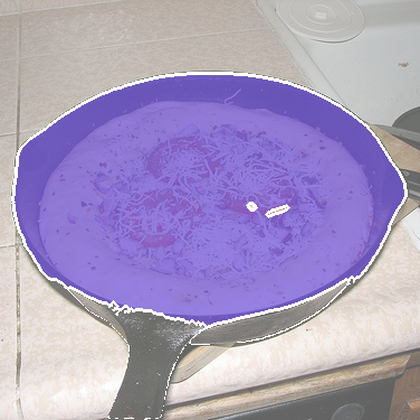}};
\node[anchor=south east, fill=white, fill opacity=0.85, text opacity=1, inner xsep=1.2pt, inner ysep=0.6pt, rounded corners=1pt, font=\sffamily\bfseries\fontsize{5.6}{6}\selectfont] at (17.330,-8.175) {0.91};
\node[rotate=90, align=center, inner sep=0, font=\sffamily\fontsize{6.2}{7}\selectfont] at (0.60,-9.047) {\textbf{racket}\\[-0.4pt]\emph{hit}};
\node[inner sep=0, anchor=south west, draw=inkgray!25, line width=0.3pt] at (0.860,-9.840) {\includegraphics[width=\fqs cm]{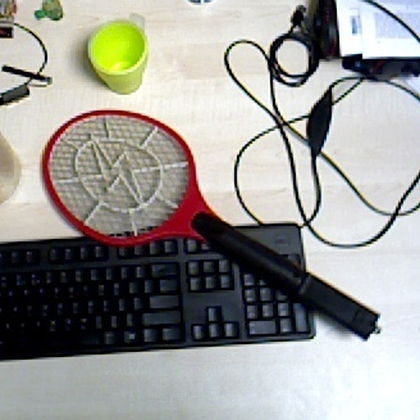}};
\node[inner sep=0, anchor=south west, draw=inkgray!25, line width=0.3pt] at (2.495,-9.840) {\includegraphics[width=\fqs cm]{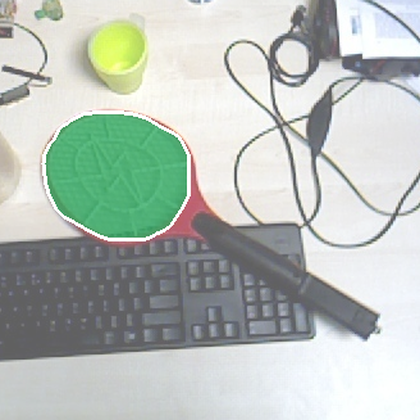}};
\node[inner sep=0, anchor=south west, draw=inkgray!25, line width=0.3pt] at (4.230,-9.840) {\includegraphics[width=\fqs cm]{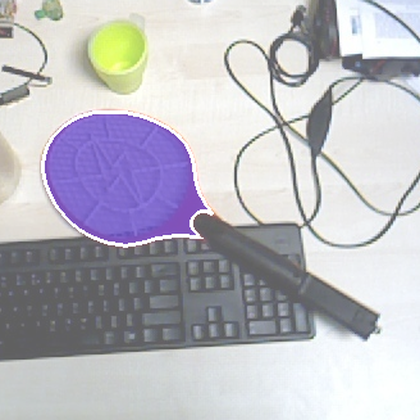}};
\node[anchor=south east, fill=white, fill opacity=0.85, text opacity=1, inner xsep=1.2pt, inner ysep=0.6pt, rounded corners=1pt, font=\sffamily\fontsize{5.6}{6}\selectfont] at (5.785,-9.810) {0.89};
\node[inner sep=0, anchor=south west, draw=inkgray!25, line width=0.3pt] at (5.865,-9.840) {\includegraphics[width=\fqs cm]{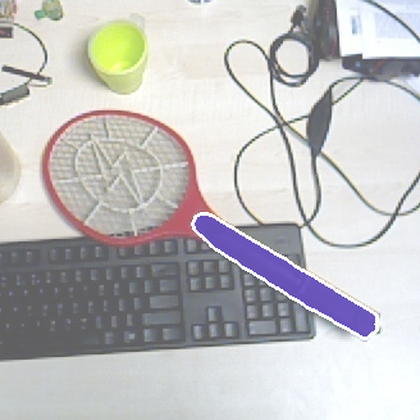}};
\node[anchor=south east, fill=white, fill opacity=0.85, text opacity=1, inner xsep=1.2pt, inner ysep=0.6pt, rounded corners=1pt, font=\sffamily\fontsize{5.6}{6}\selectfont] at (7.420,-9.810) {0.00};
\node[inner sep=0, anchor=south west, draw=inkgray!25, line width=0.3pt] at (7.500,-9.840) {\includegraphics[width=\fqs cm]{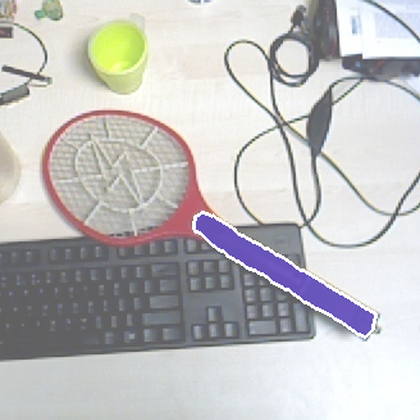}};
\node[anchor=south east, fill=white, fill opacity=0.85, text opacity=1, inner xsep=1.2pt, inner ysep=0.6pt, rounded corners=1pt, font=\sffamily\fontsize{5.6}{6}\selectfont] at (9.055,-9.810) {0.00};
\node[inner sep=0, anchor=south west, draw=inkgray!25, line width=0.3pt] at (9.135,-9.840) {\includegraphics[width=\fqs cm]{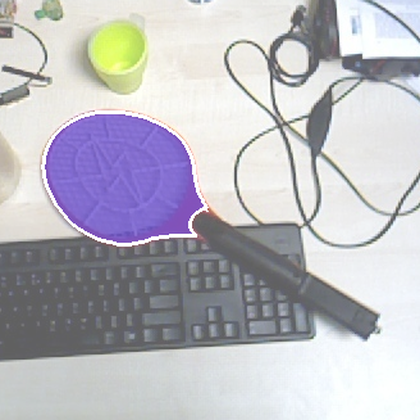}};
\node[anchor=south east, fill=white, fill opacity=0.85, text opacity=1, inner xsep=1.2pt, inner ysep=0.6pt, rounded corners=1pt, font=\sffamily\fontsize{5.6}{6}\selectfont] at (10.690,-9.810) {0.90};
\node[inner sep=0, anchor=south west, draw=inkgray!25, line width=0.3pt] at (10.770,-9.840) {\includegraphics[width=\fqs cm]{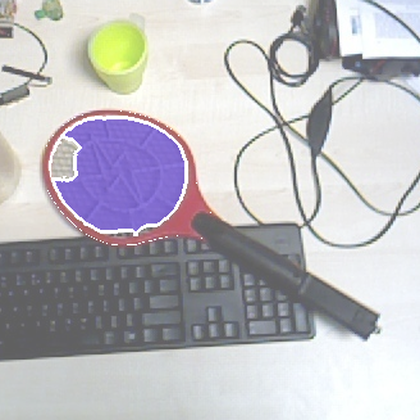}};
\node[anchor=south east, fill=white, fill opacity=0.85, text opacity=1, inner xsep=1.2pt, inner ysep=0.6pt, rounded corners=1pt, font=\sffamily\fontsize{5.6}{6}\selectfont] at (12.325,-9.810) {0.78};
\node[inner sep=0, anchor=south west, draw=inkgray!25, line width=0.3pt] at (12.405,-9.840) {\includegraphics[width=\fqs cm]{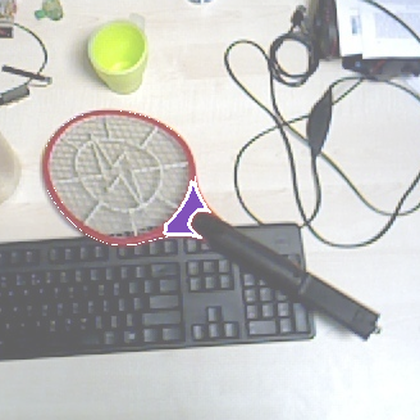}};
\node[anchor=south east, fill=white, fill opacity=0.85, text opacity=1, inner xsep=1.2pt, inner ysep=0.6pt, rounded corners=1pt, font=\sffamily\fontsize{5.6}{6}\selectfont] at (13.960,-9.810) {0.02};
\node[inner sep=0, anchor=south west, draw=inkgray!25, line width=0.3pt] at (14.040,-9.840) {\includegraphics[width=\fqs cm]{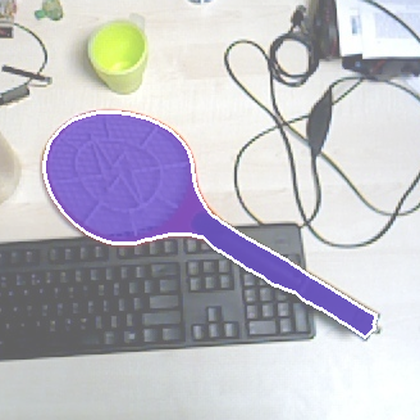}};
\node[anchor=south east, fill=white, fill opacity=0.85, text opacity=1, inner xsep=1.2pt, inner ysep=0.6pt, rounded corners=1pt, font=\sffamily\fontsize{5.6}{6}\selectfont] at (15.595,-9.810) {0.68};
\node[inner sep=0, anchor=south west, draw=cvprblue, line width=0.9pt] at (15.775,-9.840) {\includegraphics[width=\fqs cm]{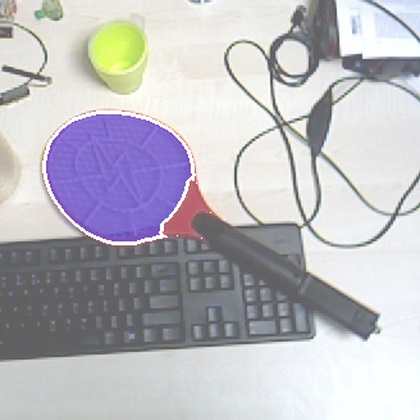}};
\node[anchor=south east, fill=white, fill opacity=0.85, text opacity=1, inner xsep=1.2pt, inner ysep=0.6pt, rounded corners=1pt, font=\sffamily\bfseries\fontsize{5.6}{6}\selectfont] at (17.330,-9.810) {0.94};
\draw[inkgray!50, line width=0.5pt] (0.30,-0.140) -- (0.30,-4.875);
\node[rotate=90, text=inkgray!85, inner sep=0, font=\sffamily\fontsize{5.8}{6.6}\selectfont] at (0.15,-2.508) {InstructPart};
\draw[inkgray!50, line width=0.5pt] (0.30,-5.045) -- (0.30,-6.510);
\node[rotate=90, text=inkgray!85, inner sep=0, font=\sffamily\fontsize{5.8}{6.6}\selectfont] at (0.15,-5.777) {UMD};
\draw[inkgray!50, line width=0.5pt] (0.30,-6.680) -- (0.30,-9.780);
\node[rotate=90, text=inkgray!85, inner sep=0, font=\sffamily\fontsize{5.8}{6.6}\selectfont] at (0.15,-8.230) {IIT-AFF};
\end{tikzpicture}
\caption{\textbf{Qualitative comparison} on official test queries. Columns: the RGB input, the ground truth (green), the seven baselines of \cref{tab:instructpart,tab:umd,tab:iit} and \method; predictions are purple and numbers are per-query IoU. Rows were picked by hand, one per object and for clear photos, among the queries on which \method reaches IoU $\geq 0.85$ and the seven baselines average at most $0.5$. Row labels give each dataset's own operation word; every method receives the same aligned words (\cref{sec:setup}).}
\label{fig:qualitative}
\end{figure*}

\smallhead{Qualitative results.}
\Cref{fig:qualitative} shows six test queries with the prediction of every method. Methods with their own inference
paths often return the most salient region of the object instead of the part that performs the operation, such as the
bowl of a wine glass for \emph{hold} or the window of an oven for \emph{open}, and the released judges, which score one
candidate per request, tend to accept scattered fragments, as on the pan. \method selects the operating part in each
row; individual baselines also succeed on some rows, such as OP-HRG on the kettle lid and the racket face and Valen on the oven handle. The two failure modes mirror the two
interfaces: a box or a few points commit to one region before the part is identified, and a judge that sees one
candidate per request has no way to prefer the whole part over a fragment of it.

\begin{table}[t]
\centering\footnotesize
\setlength{\tabcolsep}{3.2pt}
\caption{\textbf{InstructPart} official test (\%), in domain: $835$ object--operation queries on $538$ photos, each
scored against the union of the object's native part masks with that operation. The alignment of
\cref{sec:setup} touches no InstructPart query, so these are both protocols at once. The System 1
judges run in our pipeline. Our judge sees only the official training annotations.}
\label{tab:instructpart}
\begin{tabular}{@{}lrrrr@{}}\toprule
Method & gIoU$\uparrow$ & cIoU$\uparrow$ & P$_{50\text{--}95}{\uparrow}$ & P$_{50}{\uparrow}$\\\midrule
Affordance-R1~\cite{wang2026affordancer1} & 61.67 & 46.12 & 50.46 & 68.26\\
VisionReasoner~\cite{liu2026visionreasoner} & 47.90 & 36.56 & 32.13 & 48.26\\
AffordanceVLM~\cite{wu2025ragnet} & 25.89 & 20.81 & 14.26 & 23.47\\
OP-HRG~\cite{mehrab2026reasoning} & 58.78 & 49.66 & 47.16 & 63.23\\\midrule
Valen~\cite{valen2026} & 53.50 & 50.77 & 37.22 & 57.60\\
H2O-Lightning~\cite{h2o2026lightning} & 44.53 & 39.89 & 27.94 & 47.43\\
JevAny-LoRA~\cite{simplejev2026jevany} & 36.93 & 31.73 & 18.32 & 33.41\\\midrule
\textbf{\method} & \textbf{71.35} & \textbf{79.24} & \textbf{57.93} & \textbf{80.00}\\\bottomrule
\end{tabular}
\end{table}

\begin{table}[t]
\centering\footnotesize
\setlength{\tabcolsep}{3.2pt}
\caption{\textbf{UMD}, zero-shot (\%): $3{,}910$ queries on $1{,}922$ manually annotated images covering all $17$
categories, sampled in proportion to each object instance's frames, under the aligned operation words of
\cref{sec:setup}.}
\label{tab:umd}
\begin{tabular}{@{}lrrrr@{}}\toprule
Method & gIoU$\uparrow$ & cIoU$\uparrow$ & P$_{50\text{--}95}{\uparrow}$ & P$_{50}{\uparrow}$\\\midrule
Affordance-R1~\cite{wang2026affordancer1} & 56.68 & 54.19 & 35.32 & 62.07\\
VisionReasoner~\cite{liu2026visionreasoner} & 47.28 & 44.69 & 17.94 & 40.64\\
AffordanceVLM~\cite{wu2025ragnet} & 34.56 & 32.32 & 17.91 & 38.36\\
OP-HRG~\cite{mehrab2026reasoning} & 62.93 & 62.02 & 44.89 & 73.76\\\midrule
Valen~\cite{valen2026} & 63.46 & 63.10 & 45.97 & 73.86\\
H2O-Lightning~\cite{h2o2026lightning} & 59.38 & 59.12 & 41.28 & 68.57\\
JevAny-LoRA~\cite{simplejev2026jevany} & 51.17 & 47.82 & 27.56 & 51.74\\\midrule
\textbf{\method} & \textbf{70.72} & \textbf{69.85} & \textbf{51.07} & \textbf{79.95}\\\bottomrule
\end{tabular}
\end{table}

\begin{table}[t]
\centering\footnotesize
\setlength{\tabcolsep}{3.2pt}
\caption{\textbf{IIT-AFF} test, zero-shot (\%), aligned operation words: $3{,}680$ queries on $2{,}315$ images,
restricted to queries whose target pixels belong to a single object category.}
\label{tab:iit}
\begin{tabular}{@{}lrrrr@{}}\toprule
Method & gIoU$\uparrow$ & cIoU$\uparrow$ & P$_{50\text{--}95}{\uparrow}$ & P$_{50}{\uparrow}$\\\midrule
Affordance-R1~\cite{wang2026affordancer1} & 54.75 & 56.48 & 38.05 & 58.48\\
VisionReasoner~\cite{liu2026visionreasoner} & 48.91 & 50.58 & 29.58 & 49.24\\
AffordanceVLM~\cite{wu2025ragnet} & 23.56 & 27.79 & 10.10 & 21.79\\
OP-HRG~\cite{mehrab2026reasoning} & 56.00 & 54.25 & 39.24 & 61.11\\\midrule
Valen~\cite{valen2026} & 53.43 & 53.28 & 35.46 & 55.62\\
H2O-Lightning~\cite{h2o2026lightning} & 47.46 & 47.77 & 30.02 & 49.86\\
JevAny-LoRA~\cite{simplejev2026jevany} & 45.82 & 45.24 & 25.08 & 46.11\\\midrule
\textbf{\method} & \textbf{60.18} & \textbf{61.43} & \textbf{43.33} & \textbf{65.30}\\\bottomrule
\end{tabular}
\end{table}

\subsection{Ablation study}
\label{sec:ablation}

\begin{table}[t]
\centering\scriptsize
\setlength{\tabcolsep}{2.2pt}
\caption{\textbf{Ablations} on the full test sets (gIoU and cIoU, \%), with the data and rules of \method. A1--A2 change
inference only; B1--C2(b) are retrained; B3 removes the verbalizer from every level.
Bold: best per column.}
\label{tab:ablation}
\begin{tabular}{@{}lrrrrrr@{}}\toprule
 & \multicolumn{2}{c}{InstructPart} & \multicolumn{2}{c}{UMD} & \multicolumn{2}{c}{IIT-AFF}\\
\cmidrule(lr){2-3}\cmidrule(lr){4-5}\cmidrule(l){6-7}
Variant & gIoU & cIoU & gIoU & cIoU & gIoU & cIoU\\\midrule
\textbf{\method} & \textbf{71.35} & \textbf{79.24} &  \textbf{70.72} & \textbf{69.85} & \textbf{60.18} & \textbf{61.43}\\
\midrule
\multicolumn{7}{@{}l}{\textit{Framework}}\\
A1 w/o hierarchy & 43.65 & 65.88 & 3.53 & 11.19 & 35.79 & 57.00\\
A2 w/o evidence pass. & 66.43 & 77.07 & 70.20 & 69.00 & 59.59 & 60.83\\
\midrule
\multicolumn{7}{@{}l}{\textit{Judge architecture}}\\
B1 Drawn markers & 67.81 & 74.25 & 63.13 & 60.19 & 53.46 & 60.33\\
B2 Per-candidate & 69.78 & 76.69 & 67.37 & 66.13 & 59.40 & 60.02\\
B3 w/o verbalizer & 71.31 & 78.84 & 68.69 & 67.65 & 58.89 & 59.76\\
\midrule
\multicolumn{7}{@{}l}{\textit{Judge training}}\\
C1 GT parts & 69.77 & 78.05 & 68.73 & 68.94 & 59.19 & 59.49\\
C2(a) Membership only & 69.54 & 71.70 & 68.43 & 67.76 & 59.74 & 59.72\\
C2(b) Purity only & 70.74 & 78.46 & 69.89 & 69.07 & 59.59 & 59.62\\
\bottomrule
\end{tabular}
\end{table}

We report gIoU and cIoU on all three test sets; intervals are paired bootstraps over photos or, on UMD, object
instances.

\smallhead{Framework.}
Both framework ablations change only inference. Without evidence passing each level is decided once, top-down,
with no backtracking or sub-part refinement: InstructPart gIoU falls from $71.35$ to $66.43$ ($[3.42, 6.52]$) and
cIoU by $2.17$.
Without the hierarchy, PartGLEE proposes parts on the whole image and the judge answers a single part question:
InstructPart gIoU falls to $43.65$, and $319$ of the $835$ queries end with no accepted region. On UMD, whose small tools
sit in clutter, it falls to $3.53$. The hierarchy is therefore the largest single factor: a single question over the
whole image confronts the judge with the parts of every object at once, and without an object level a wrong object
cannot be corrected. Evidence passing then adds what the levels alone cannot, mostly through sub-parts that rescue a
rejected part or trim an accepted one.

\smallhead{Judge architecture and training.}
B2 judges one candidate per request at equal compute. Drawing numbered markers onto the image instead of region tokens costs $3.54$ gIoU ($[1.95, 5.21]$) and $4.99$
cIoU in domain, and $7.59$ and $6.72$ gIoU on UMD and IIT-AFF: a painted number changes the pixels the judge sees and
leaves a small part to be located from its mark, whereas a region token is built from the part's own features. One candidate per request costs $1.57$ gIoU and $2.55$ cIoU in domain, and $3.35$ and $0.78$ gIoU on UMD and IIT-AFF. Removing the verbalizer costs $2.03$ gIoU on UMD
($[0.91, 3.17]$) and $1.29$ on IIT-AFF ($[0.83, 1.76]$), and cIoU drops on both. Training on clean ground-truth parts in place of the
framework's candidates costs $1.58$ gIoU in domain, $1.99$ on UMD and $0.99$ on IIT-AFF, consistent with training
the judge on the candidates it will see (\cref{sec:training}).
Training only the membership output costs $1.81$ gIoU ($[0.66, 3.05]$) and $7.54$ cIoU; training only the purity
output costs $0.61$ gIoU and $0.78$ cIoU in domain. The two outputs are complementary: membership alone costs far more
cIoU than gIoU, since cIoU pools pixels and so penalises accepted regions that extend beyond the target, and the full judge with
both outputs is best on every column.

\subsection{Analysis}
\label{sec:analysis}
\smallhead{Error analysis.}
The candidates admit far more than any judge extracts: accepting exactly the candidates and sub-parts that lie
mostly inside the ground truth would reach $78.8$, $80.6$ and $74.8$ gIoU on InstructPart, UMD and IIT-AFF, against
$71.4$, $70.7$ and $60.2$ for \method. A judge that accepted exactly these candidates would gain more than seven
points in domain without any change to the segmenters, which makes the judge the natural place for further
improvement.

\smallhead{Definition alignment.}
The alignment changes only the words of the conflicting classes. On the earlier UMD set, where both protocols were
run, the three operations it leaves untouched --- \emph{cut}, \emph{scoop} and \emph{pound} --- score identically
under both.
Results under the original words on the earlier sets are in the supplementary material.

\section{Conclusion}
\label{sec:conclusion}
We ground affordances with a System 1 model that verifies candidate regions instead of reasoning in text. The
framework asks a hierarchy of region questions --- object, part, sub-part --- and passes evidence between the levels
so that they correct one another; the model is trained for these questions from raw weights on the framework's own
requests and judges each level in one pass. Under definition-aligned evaluation on the full InstructPart, UMD and
IIT-AFF test sets it leads every published method on every column, with about a quarter of the judge requests of
per-candidate judges. Beyond affordances, the same recipe
--- frozen proposals, a fixed hierarchy of region questions and a judge that answers each in one pass --- applies to
other part-level grounding tasks, and to robotic settings in which a reasoning trace is too slow to produce at every
step. More broadly, the results show that for affordance grounding the structure of the questions can carry the
reasoning that chain-of-thought methods write out, while improving accuracy in domain and in transfer.

\section*{LLM Usage Statement}
We used OpenAI's GPT and Anthropic's Claude at the \emph{Assist} level to edit or refine content and materials
written, designed or produced by the authors, covering paper writing, related work and citations, and code. In all
categories, the involvement of the AI tools was limited to refining the authors' existing work. The authors take full
responsibility for the accuracy of the submitted work, including its implementation, experiments and analysis of
results, figures and visual materials, and data preparation and annotation, and for the academic integrity and
reproducibility of its research ideas, hypotheses, claims and conclusions.

{\small
\bibliographystyle{ieeenat_fullname}
\bibliography{references}
}
\end{document}